\documentclass{article}

\PassOptionsToPackage{numbers,sort&compress}{natbib}
\usepackage[main,final]{neurips_2026}

\usepackage[utf8]{inputenc}
\usepackage[T1]{fontenc}
\usepackage{hyperref}
\hypersetup{hidelinks,pdftitle={Feeling of Knowing in Large Language Models},pdfauthor={Zichuan Fu, Xian Wu, Jingtong Gao, Wenlin Zhang, Jiaxuan Li, Binhao Wang, Yimin Deng, Guojing Li, Xiaopeng Li, Derong Xu, Yefeng Zheng, Xiangyu Zhao}}
\usepackage{url}
\usepackage{booktabs}
\usepackage{array}
\usepackage{amsfonts}
\usepackage{amsmath}
\usepackage{amssymb}
\usepackage{amsthm}
\usepackage{nicefrac}
\usepackage{microtype}
\usepackage{xcolor}
\usepackage[normalem]{ulem}
\definecolor{takeawayback}{RGB}{255,248,235}
\definecolor{takeawayframe}{RGB}{224,132,39}
\definecolor{opsftdelete}{RGB}{255,232,232}
\definecolor{opsftinsert}{RGB}{226,246,232}
\newcommand{\delspan}[1]{\begingroup\setlength{\fboxsep}{1pt}\colorbox{opsftdelete}{\sout{#1}}\endgroup}
\newcommand{\addspan}[1]{\begingroup\setlength{\fboxsep}{1pt}\colorbox{opsftinsert}{#1}\endgroup}
\usepackage{tcolorbox}
\newtcolorbox{takeawaybox}{colback=takeawayback,colframe=takeawayframe,
  top=1mm,bottom=1mm,left=1.5mm,right=1.5mm,boxrule=0.5pt,arc=1mm}
\usepackage{graphicx}
\usepackage{subcaption}
\usepackage{wrapfig}
\usepackage{enumitem}
\usepackage{flafter}
\usepackage{placeins}
\usepackage{algorithm}
\usepackage{algorithmic}
\usepackage{multirow}

\newcolumntype{M}{>{\centering\arraybackslash}c}
\newcolumntype{N}{>{\centering\arraybackslash}p{1.0cm}}

\title{Feeling of Knowing in Large Language Models}

\author{%
  \textbf{Zichuan Fu\textsuperscript{1,}\thanks{Work done during an internship at Tencent Jarvis Lab.}\hphantom{$^{*}$}, Xian Wu\textsuperscript{2,$\dagger$}, Jingtong Gao\textsuperscript{1,$\dagger$}, Wenlin Zhang\textsuperscript{1}, Jiaxuan Li\textsuperscript{1}, Binhao Wang\textsuperscript{1},}\\
  \textbf{Yimin Deng\textsuperscript{1,3}, Guojing Li\textsuperscript{1,4}, Xiaopeng Li\textsuperscript{1}, Derong Xu\textsuperscript{1,5}, Yefeng Zheng\textsuperscript{6}, Xiangyu Zhao\textsuperscript{1,}\thanks{Corresponding authors.}}\\
  \normalfont\textsuperscript{1}City University of Hong Kong, \textsuperscript{2}Tencent Jarvis Lab\\
  \normalfont\textsuperscript{3}Xi'an Jiaotong University, \textsuperscript{4}Renmin University of China\\
  \normalfont\textsuperscript{5}University of Science and Technology of China, \textsuperscript{6}Westlake University\\
  \normalfont\href{mailto:zc.fu@my.cityu.edu.hk}{\texttt{zc.fu@my.cityu.edu.hk}},
  \href{mailto:kevinxwu@tencent.com}{\texttt{kevinxwu@tencent.com}},\\
  \normalfont\href{mailto:jt.g@my.cityu.edu.hk}{\texttt{jt.g@my.cityu.edu.hk}},
  \href{mailto:xianzhao@cityu.edu.hk}{\texttt{xianzhao@cityu.edu.hk}}
}

\begin{document}

\maketitle

\begin{abstract}
Large language models (LLMs) are frequently updated through post-training techniques to maintain up-to-date knowledge in real-world deployment. Their reliability depends not only on what the LLM knows, but also on whether it knows what it knows---which we study as functional \emph{feeling of knowing} (FoK).
FoK serves as the underlying signal for selective generation, retrieval triggering, and model routing. Miscalibrated FoK makes systems overconfident on unknown questions or unnecessarily refuse questions they could have answered.
As LLMs are updated through post-training, however, we discover that what an LLM knows can change without a corresponding update to its FoK. This creates a knowledge--FoK desynchronization challenge, where FoK estimators calibrated to the pre-trained LLM become unreliable as the model's knowledge state evolves.
To address this challenge, we propose a two-channel design: a knowledge channel that changes what
the LLM knows and a FoK channel that judges whether the current knowledge
state supports answering. We train these channels with a three-stage
procedure: on-policy supervised fine-tuning (OP-SFT) updates the knowledge
channel from corrected on-policy answers, knowledge-delta FoK training
(K$\Delta$-FT) trains the FoK channel over intermediate knowledge states,
and FoK-guided policy optimization (FGPO) uses the judgment to generate
the corresponding answer or explanation. Experiments on four datasets show that explicit FoK alignment substantially
improves judgment quality on updated knowledge. The code is available at
\href{https://github.com/Applied-Machine-Learning-Lab/NeurIPS2026_Know-thyself}{\nolinkurl{Applied-Machine-Learning-Lab/NeurIPS2026_Know-thyself}}.
\end{abstract}

\section{Introduction}
\label{sec:intro}

Large language models (LLMs) serve as assistants, tool-using and search agents \citep{ren2023,yang2024,li2025aisearch,zhang2025deepresearch}, and recommendation and decision-support models \citep{fu2025unictr,liu2025llm4cdsr,liu2024leader}. Their reliability depends on answer accuracy and on judging whether to answer or seek external support. Cognitive science calls this metacognitive judgment \emph{feeling of knowing} (FoK) \citep{hart1965,koriat1993}. We use \emph{functional FoK} for this behavior in LLMs: answerability judgments modeled after human FoK. This scalar judgment supports selective prediction, model routing, and retrieval triggering \citep{kadavath2022,fu2026tandem,xu2025amar,jia2024mill,li2025personalization}. If it fails to reflect updated knowledge, a system may confidently answer unknown questions or refuse questions the updated model could solve.

Existing LLM FoK methods mainly take three forms. Output-based
methods read uncertainty from the model's own generations, such as
the probability of a ``true'' token or the entropy of sampled answers
\citep{kadavath2022,kuhn2023,farquhar2024}. Probe-based methods train
a lightweight classifier on hidden states
\citep{azaria2023,burns2022}. Elicitation-based methods supervise the
model to verbalize its confidence directly
\citep{tian2023,xiong2024,elical2025}. However, these methods are
usually calibrated and evaluated under static model parameters. This
leaves open what happens when the model is later updated by
supervised fine-tuning (SFT), low-rank adaptation (LoRA), knowledge
editing, or continual learning \citep{meng2022rome,fu2025mmke,jang2022,luo2023}.
After an update, the LLM's knowledge may change, while a FoK estimator trained or calibrated for the base model may still judge answerability based on the base model's knowledge boundary.
\textbf{The problem of keeping FoK aligned under changing model
knowledge remains underexplored.}

Koriat's account motivates this study: human FoK is inferred from accessible cues rather than directly read from memory \citep{koriat1993}. Our correctness-supervised objective does not instantiate or validate that cognitive process. In LLMs, prior causal tracing locates factual associations mainly in middle MLP layers \citep{meng2022rome}, whereas our FoK recovery peaks occur later (Section~\ref{subsec:f1}). Two challenges follow: 1) \textbf{Knowledge--FoK desynchronization:} knowledge training changes answerability without keeping the FoK judgment aligned; and 2) \textbf{Context-dependent FoK:} the judgment varies with the available context and generated reasoning. We therefore train a FoK channel to judge the current knowledge state and to guide the subsequent response.

We address these challenges with two trainable channels. On-policy supervised fine-tuning (OP-SFT) updates the knowledge channel from the model's own answers, supervising only corrected answer-relevant suffixes to limit disruption to FoK. Knowledge-delta FoK training (K$\Delta$-FT) trains the FoK channel across knowledge deltas, allowing judgment labels to reflect changing answerability. For context-dependent FoK, FoK-guided policy optimization (FGPO) trains the model to generate context, judge answerability, and then answer or explain.
The paper makes three contributions:
\begin{itemize}[leftmargin=*]
  \item We study functional FoK after knowledge updates, finding later-layer FoK recovery, sensitivity to accessible context, and that knowledge updating alone fails to synchronize answerability judgments.
  \item We introduce a two-channel framework with OP-SFT, K$\Delta$-FT, and FGPO to update knowledge, synchronize FoK, and train the model to answer or explain according to its FoK judgment.
  \item We evaluate FoK on factual QA with cross-dataset transfer, three sequential updates, and seed variation for the main method. MMLU and ARC-Easy measure capability preservation.
\end{itemize}

\section{Mechanism of LLM FoK}
\label{sec:findings}
\label{sec:problem}

\subsection{Problem formulation}
\label{subsec:formal-setting}

Let $M_\theta$ be an LLM at parameter state $\theta$, and let $q$ be a
question. We use $Y_\theta(q)\in\{0,1\}$ for whether the model's greedy
answer is correct, and $J_\theta(q)\in[0,1]$ for the model's
functional FoK score, where 0 means the model judges that it
cannot answer and 1 means it judges that it can answer. The quality
of FoK is reflected in the alignment between $J_\theta(q)$ and
$Y_\theta(q)$ under the same decoding rule. For a generative yes/no
judgment prompt, we instantiate this score as
$J_\theta(q)=
\frac{P_\theta(\mathrm{yes}\mid q)}
{P_\theta(\mathrm{yes}\mid q)+P_\theta(\mathrm{no}\mid q)}$.

When an LLM learns new knowledge through post-training, its parameters
change from the base state $\theta$ to an updated state
$\tilde{\theta}$. We call the induced difference
$\kappa=\tilde{\theta}-\theta$ the \emph{knowledge delta}. The
target is to keep $J_{\tilde{\theta}}$ aligned with
$Y_{\tilde{\theta}}$ as knowledge changes. Our method trains
FoK along the update trajectory.

In evaluation, we focus on two directional failures of the FoK
judgment. An \emph{illusion} occurs when the model
judges that it can answer but is wrong; an \emph{omission} occurs
when it judges that it cannot answer although its answer would be
correct. Metrics and splits are detailed in
Section~\ref{subsec:exp-settings}.

\newcommand{\gain}[1]{\textcolor{green!45!black}{#1}}
\newcommand{\loss}[1]{\textcolor{red!70!black}{#1}}

\begin{figure}[t]
  \centering
  \begin{minipage}[t]{0.48\linewidth}
    \centering
    \vspace{0pt}
    \includegraphics[width=\linewidth]{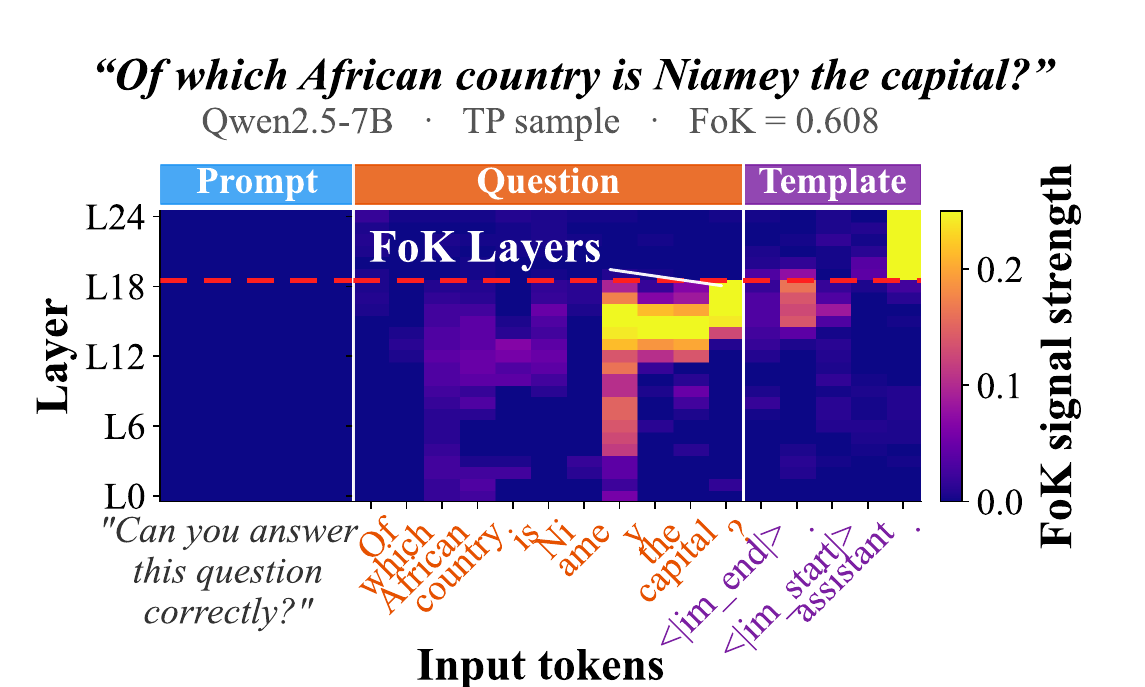}
    \subcaption{Causal tracing of FoK signal in a 7B LLM.}
    \label{fig:f1-tp-single}
  \end{minipage}\hfill%
  \begin{minipage}[t]{0.48\linewidth}
    \centering
    \vspace{0pt}
    \includegraphics[width=\linewidth]{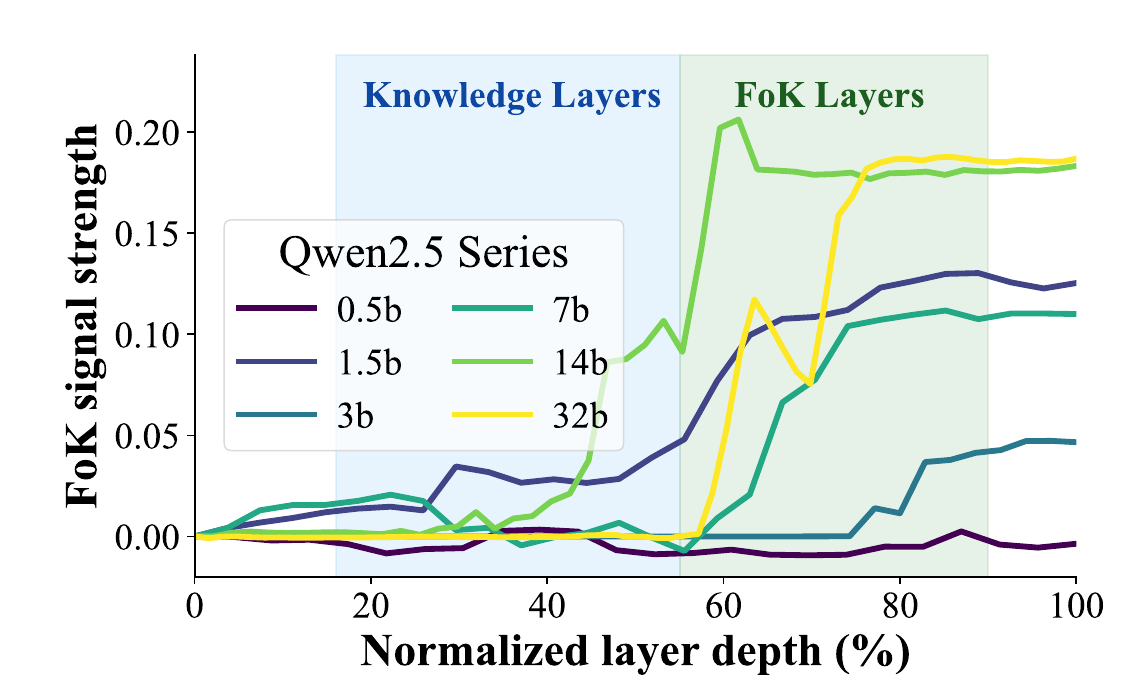}
    \subcaption{FoK signal across normalized depth.}
    \label{fig:f1-recovery}
  \end{minipage}
  \caption{FoK signal localizes late under causal restoration. Panel (a)
  shows a 7B example across tokens and layers; panel (b) aggregates FoK
  recovery across Qwen2.5 sizes over normalized depth.}
  \label{fig:f1-main}
\end{figure}

\begin{table}[t]
  \caption{Knowledge and FoK under separate or sequential SFT updates.
  Base gives absolute values; the three SFT blocks give changes from Base.
  Accuracy changes are percentage points (pp); AUROC changes are unitless.
  $\Delta$AUC denotes the change in AUROC. Higher accuracy and AUROC are better. Full illusion and omission rates
  are reported in Appendix~\ref{app:detailed-preliminary}.}
  \label{tab:f2-knowledge-judgment-coupling}
  \centering
  \begin{tabular}{@{}lcccccccc@{}}
    \toprule
    & \multicolumn{2}{c}{Base} & \multicolumn{2}{c}{K-SFT}
    & \multicolumn{2}{c}{J-SFT} & \multicolumn{2}{c}{K$\to$J SFT}\\
    \cmidrule(lr){2-3}\cmidrule(lr){4-5}\cmidrule(lr){6-7}\cmidrule(l){8-9}
    Model & Acc. & AUROC & $\Delta$Acc. & $\Delta$AUC
    & $\Delta$Acc. & $\Delta$AUC & $\Delta$Acc. & $\Delta$AUC\\
    \midrule
    Qwen2.5-0.5B & 20\% & 0.519 & +2 & -0.016 & -14 & +0.213 & +2 & +0.211 \\
    Qwen2.5-1.5B & 35\% & 0.529 & +6 & -0.035 & +2 & +0.255 & +5 & +0.248 \\
    Qwen2.5-3B & 39\% & 0.739 & +12 & -0.220 & +3 & +0.038 & +11 & +0.051 \\
    Qwen2.5-7B & 53\% & 0.724 & +7 & -0.042 & +2 & +0.088 & +7 & +0.100 \\
    Qwen2.5-14B & 66\% & 0.762 & +6 & +0.004 & +1 & +0.030 & +5 & +0.068 \\
    Qwen2.5-32B & 68\% & 0.819 & +4 & -0.026 & 0 & +0.010 & +4 & +0.020 \\
    LLaMA-3.1-8B & 74\% & 0.672 & +2 & -0.038 & +1 & +0.045 & +2 & +0.063 \\
    \bottomrule
  \end{tabular}
\end{table}

\subsection{FoK localization in model depth}
\label{subsec:f1}

To localize FoK ability, we first adopt ROME's causal-tracing
method \citep{meng2022rome}. For each factual question, we corrupt the
subject token, restore hidden states one layer at a time, and measure
whether the yes/no FoK judgment is recovered. We repeat the procedure
across Qwen2.5 model sizes, using normalized depth to compare
corresponding layers across architectures rather than raw layer indices.

As shown in Figure~\ref{fig:f1-tp-single}, Qwen2.5-7B places
the strongest FoK signal in late layers, around two-thirds of model depth.
Figure~\ref{fig:f1-recovery} shows that the same late-layer pattern
remains after normalizing depth across Qwen2.5 scales from 0.5B to 32B.
This is later than the factual association signal reported by ROME,
which is localized in middle MLP layers, around one-third of model depth
\citep{meng2022rome}. Appendix Figure~\ref{fig:f1-weight-delta} gives a
parameter-side check: knowledge SFT changes shallow-to-middle layers
more, while judgment SFT changes concentrate in the late layers where
FoK signal strength is high. Appendix~\ref{app:f1-causal-tracing}
extends the causal tracing across scales and outcome groups.

\begin{takeawaybox}
\textbf{Takeaway 1}: Under causal tracing, the FoK signal peaks later in
model depth than factual associations, suggesting that FoK acts as a
judgment over the model's knowledge state rather than as factual recall
itself.
\end{takeawaybox}

\subsection{How FoK depends on updated knowledge}
\label{subsec:f2}

Depth separation does not mean independence. To analyze how
factual knowledge and FoK relate, we
intervene on one side and evaluate the other under matched data and
optimization settings. First, we update factual knowledge with vanilla
factual SFT and measure FoK on the updated model. Second, we train FoK
alone and measure whether answer accuracy changes. Table~\ref{tab:f2-knowledge-judgment-coupling}
summarizes cross-scale results. Appendix~\ref{app:full_table} adds exact-match (EM) results
across scales and families.

Table~\ref{tab:f2-knowledge-judgment-coupling} shows that knowledge
SFT (K-SFT) increases answer accuracy for every model, but FoK AUROC
decreases for six of the seven models, with the largest drop on
Qwen2.5-3B. Thus, the knowledge update changes the model's answerability,
but the judgment boundary does not adjust correctly. The mismatch is
not a single error direction: after K-SFT, some models make more
omissions, judging that they cannot answer questions they would answer
correctly, whereas larger Qwen models better preserve AUROC but
shift toward illusion, judging that they can answer even when their
answers are wrong. The issue is therefore not simply increased
conservatism or confidence, but a weakened alignment between the
current knowledge state and FoK. Judgment SFT (J-SFT) provides a
reference intervention: directly training judgment improves FoK
without compromising answer performance in larger models. K$\to$J
SFT further shows that judgment training after a knowledge update can
recalibrate FoK while preserving most of the answer-accuracy gain from
K-SFT, indicating that FoK depends on the current knowledge state and
should be updated after knowledge changes. Appendix~\ref{app:f2-supplementary}
provides the FoK-recovery and reverse-direction results.

\begin{takeawaybox}
\textbf{Takeaway 2}: In current LLMs, knowledge updates change
answerability but do not automatically update FoK. FoK depends on
features of the current knowledge state, so changing knowledge can
break the mapping from knowledge to self-judgment unless
FoK is explicitly updated.
\end{takeawaybox}

\subsection{Reasoning before judgment increases FoK certainty}
\label{subsec:f3}

\begin{figure}[t]
  \centering
  \includegraphics[width=0.52\linewidth]{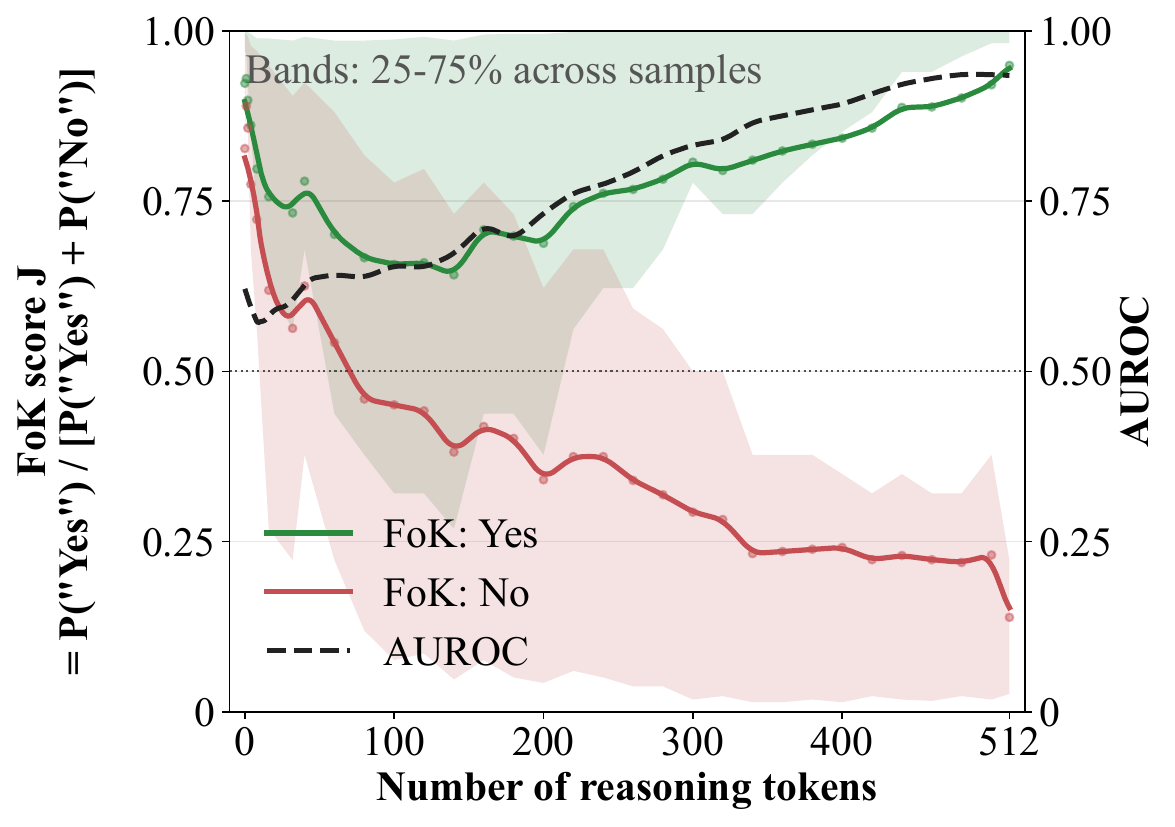}
  \caption{FoK score and AUROC over pre-judgment reasoning tokens. Green/red
  curves show mean FoK score for examples whose final FoK token is Yes/No.
  $J=0.5$ denotes FoK uncertainty, while $J\approx1$ or $J\approx0$ denotes
  high certainty. The black curve reports AUROC of prefix-level $J$ for
  predicting whether the final FoK token is Yes or No.}
  \label{fig:f3-reasoning-tokens}
\end{figure}

Koriat's FoK theory suggests that the amount and intensity of
accessible information increase FoK judgments \citep{koriat1993}. We
test the LLM analogue by asking whether self-generated reasoning
affects FoK.

We evaluate Qwen2.5-7B on MATH~\citep{hendrycks2021math}
problems. For each problem, the model first generates a solution trace.
We then re-query the same model at increasing prefixes of that trace
and compute
$J=P(\textrm{``Yes''})/[P(\textrm{``Yes''})+P(\textrm{``No''})]$ for
the question ``Can you solve this problem correctly?''
Figure~\ref{fig:f3-reasoning-tokens} shows that the FoK signal becomes
more separated over the generated trace. As more reasoning tokens are
available, the mean FoK scores for final ``Yes'' and ``No'' judgments
move toward $J=1$ and $J=0$, respectively; the AUROC of $J$ for the
final Yes/No FoK token rises as more reasoning tokens are available.
This indicates that FoK is sensitive to the model's current context, not
only to the original question.

\begin{takeawaybox}
\textbf{Takeaway 3}: FoK should be judged from the model's current
context, not from the original question alone. This motivates a
response-generation procedure in which the model first accumulates
task-relevant context, then evaluates answerability before producing
the final answer.
\end{takeawaybox}

\section{Methodology}
\label{sec:method}

Our methodology aligns FoK with updated knowledge through a
two-channel model structure and a three-stage procedure that updates knowledge, synchronizes FoK, and trains FoK-guided responses.

\subsection{Framework overview}
\label{subsec:framework-overview}

Starting from a base LLM $M$, we represent post-training changes as parameter
deltas \citep{fu2025tfmerge,fu2025mmke,zhang2023promptst,fu2025unictr}. The knowledge delta $\kappa$ forms the knowledge-retrieval
channel, which changes what the model knows and retrieves. The
FoK delta $\varphi$ forms the FoK channel, which judges whether the
current knowledge state supports a correct answer. 
The training procedure has three parts:
\begin{itemize}[leftmargin=*]
  \item \textbf{On-policy supervised fine-tuning (OP-SFT).}
  OP-SFT updates $\kappa$ on the model's own sampled responses and supervises
  only the corrected answer-relevant suffix. This part learns the target facts while
  saving a trajectory of knowledge states
  $\{\kappa_0,\kappa_1,\ldots,\kappa_T\}$ that later serves as
  supervision for FoK adaptation under the same base model.

  \item \textbf{Knowledge-delta FoK training (K$\Delta$-FT).} K$\Delta$-FT trains $\varphi$ over the OP-SFT checkpoint trajectory $\kappa_t$. This lets FoK judgments follow the model's knowledge as it evolves, rather than remaining tied to a fixed knowledge boundary.

  \item \textbf{FoK-guided policy optimization (FGPO).}
  Finally, FGPO trains the model to generate a reasoning trace, make a
  yes/no FoK judgment, and produce either an answer or an explanation of
  why it cannot answer. Its reward evaluates the judgment and response
  together, so FoK is optimized as part of the generated decision rather
  than as a separate post-processing score.
\end{itemize}
With this two-channel update, FoK guides the LLM before answering. The model can first judge whether its current knowledge is sufficient, then either generate an answer or explain what knowledge is missing, which can support routing to a stronger model or retrieval-augmented generation.

\subsection{On-policy supervised fine-tuning (OP-SFT)}
\label{subsec:stage1}

Section~\ref{subsec:f2} shows that standard knowledge training does not synchronize FoK with the updated knowledge state, and can severely interfere with the model's original FoK ability. To reduce this interference, we introduce on-policy supervised fine-tuning (OP-SFT). Specifically, OP-SFT keeps the model's own generated context and applies gradients only to the corrected answer-relevant span.

For each question $q$ with gold answer set $\mathcal{A}_q$, we collect
$k$ complete sampled responses from the current LLM. For a sampled response
$r=(r_{<s}, r_{[s:e]}, r_{>e})$, $r_{[s:e]}$ denotes the generated answer
span. We then handle two cases. If the sampled response already answers correctly,
we keep this answer span. If it answers incorrectly, we replace only this
answer span with a gold answer $a^*(q)\in\mathcal{A}_q$ and leave the
rest of the generated response unchanged. The repaired training response is
\begin{equation}
  \bar r(q,r) = \bigl(r_{<s}, \bar a(q,r), r_{>e}\bigr),
  \qquad
  \bar a(q,r)=
  \begin{cases}
    r_{[s:e]} & r \text{ answers correctly},\\
    a^*(q)    & r \text{ answers incorrectly}.
  \end{cases}
\end{equation}
The training sequence is therefore not a static QA pair:
it is sampled from the current LLM, changes only the wrong answer span, and uses loss only on $\bar a(q,r)$.
For example, for $q=$``Who wrote \emph{Pride and Prejudice}?'' and
$\mathcal{A}_q=\{\text{Jane Austen}\}$, OP-SFT repairs one sampled
response as
\begin{equation}
\label{eq:opsft-repair-example}
\begin{aligned}
q &: \text{Who wrote \emph{Pride and Prejudice}?},
\qquad
\mathcal{A}_q=\{\text{Jane Austen}\},\\
\bar r(q,r) &: \text{The book is a nineteenth-century novel. I think the author was}\\
&\quad \text{\delspan{Charlotte Bront\"e} \addspan{Jane Austen}, because she wrote English classics.}
\end{aligned}
\end{equation}
The red span is the first answer-relevant error in the sampled response;
the green span is the supervised correction. The remaining generated text
stays in the response, but receives no loss.
The OP-SFT objective is therefore the local corrected-span loss:
\begin{equation}
  \mathcal{L}_{\mathrm{OP\text{-}SFT}}
  = -\,\mathbb{E}_{q,r}\sum_{u=1}^{|\bar a(q,r)|}
  \log P_{M_{\kappa}}
  \bigl(\bar a(q,r)_u \mid q,r_{<s},\bar a(q,r)_{<u}\bigr).
\end{equation}
We save checkpoints $\{\kappa_0,\kappa_1,\ldots,\kappa_T\}$ during
OP-SFT, keeping a sequence of knowledge states.

\subsection{\texorpdfstring{Knowledge-delta FoK training (K$\Delta$-FT)}{Knowledge-delta FoK training (KDelta-FT)}}
\label{subsec:stage2}

After a knowledge update, the LLM's answerability changes, but its FoK does not automatically synchronize with that change.
Training FoK only on the base or final LLM gives a state-specific
answerability boundary that generalizes poorly across knowledge deltas.
Final-state training also creates label imbalance because many updated
questions become answerable.

To make FoK generalize across knowledge deltas, we introduce
knowledge-delta FoK training (K$\Delta$-FT), which uses the
OP-SFT checkpoint trajectory $\{\kappa_0,\kappa_1,\ldots,\kappa_T\}$
as supervision states rather than supervising only one endpoint \citep{zhou2025gdaug}. For each
checkpoint $\kappa_t$, we combine it with the same FoK delta $\varphi$ and
train $\varphi$ to predict whether $M_{\kappa_t}$ can answer question $q$
with the following loss:
\begin{equation}
\label{eq:kdelta-state-loss}
  \ell_{\varphi}(q,\kappa_t)
  =
  -\log P_{M_{\kappa_t,\varphi}}
  \left(
    \mathbf{1}\!\left[
      \mathrm{ans}(M_{\kappa_t},q)\in\mathcal{A}_q
    \right]
    \mid T_J(q)
  \right).
\end{equation}
Here $\mathbf{1}[\cdot]$ denotes the indicator function: it equals 1
when the answer generated by $M_{\kappa_t}$ belongs to $\mathcal{A}_q$,
and equals 0 otherwise. $T_J(q)$ wraps $q$ in the fixed Yes/No prompt
``Can you answer this question correctly?'' $M_{\kappa_t,\varphi}$ is
the base model with the frozen knowledge delta and trainable FoK delta.
For $y\in\{0,1\}$, $P_{M_{\kappa_t,\varphi}}(y\mid T_J(q))$ is the
normalized Yes/No probability: $J(q)$ for $y=1$ and $1-J(q)$ for $y=0$.
The probability is evaluated under
$M_{\kappa_t,\varphi}$, so FoK is trained with the same combined
knowledge-and-FoK state used for judgment. The overall objective
averages this checkpoint-specific loss over all questions and all
OP-SFT checkpoints:
\begin{equation}
\label{eq:kdelta-objective}
  \mathcal{L}_{\mathrm{K}\Delta\mathrm{\text{-}FT}}(\varphi)
  =
  \frac{1}{|\mathcal{Q}|(T+1)}
  \sum_{q\in\mathcal{Q}}\sum_{t=0}^{T}
  \ell_{\varphi}(q,\kappa_t).
\end{equation}
Only $\varphi$ is optimized in this stage; the knowledge checkpoints are
fixed supervision states. Because the same question is evaluated under
multiple $\kappa_t$, K$\Delta$-FT trains FoK to follow answerability as
it changes from pre-update to post-update knowledge, rather than to
memorize a static question-level label detached from the current model
state during updating.

\subsection{FoK-guided policy optimization (FGPO)}
\label{subsec:stage3}

After OP-SFT and K$\Delta$-FT, a continuous FoK score still requires a decision threshold. FoK-guided policy optimization (FGPO) trains the judgment and subsequent response together, allowing the model to commit to answering or explaining within one generation \citep{zhao2018pagewise,zhao2018deers,zhao2021usersim,liu2023latent}. It generates a short context trace, makes a FoK judgment, and produces an answer or explanation:
\begin{center}\small
\begin{tabular}{@{}c@{}}
\texttt{<think>...reasoning...</think> <answer>yes|no</answer>}\\
\texttt{<response>answer or explanation</response>}
\end{tabular}
\end{center}
For a sampled trajectory $z=(h,j,u)$, where $h$ is the reasoning trace,
$j\in\{\mathrm{yes},\mathrm{no}\}$ is the FoK judgment, and $u$ is the
response, we decompose the reward into FoK, answer, and format terms:
\begin{equation}
  R(q,z)
  =
  R_{\mathrm{FoK}}\!\left(Y(q),j\right)
  +
  R_{\mathrm{ans}}\!\left(q,Y(q),j,u\right)
  +
  R_{\mathrm{fmt}}(z)
  -
  \lambda |h|.
\end{equation}
Here $Y(q)$ is the answerability target for the current training
example. Let $C(q,u)=\mathbf{1}[u\in\mathcal{A}_q]$ denote answer
correctness, $E(q,u)$ denote whether an LLM judge accepts $u$ as a valid
explanation for not answering, and $F(z),S(z)\in\{0,1\}$ denote valid
format and proper sequence termination. The FoK term directly supervises
the judgment token:
\begin{equation}
  R_{\mathrm{FoK}}(Y,j)
  =
  \alpha_{+}\mathbf{1}[\iota(j)=Y]
  -
  \alpha_{-}\mathbf{1}[\iota(j)\neq Y],
\end{equation}
where $\iota(\mathrm{yes})=1$ and $\iota(\mathrm{no})=0$. The answer term scores the response under the
chosen judgment:
\begin{align}
R_{\mathrm{ans}}(q,Y,j,u)
=&\ \mathbf{1}[j=\mathrm{yes},Y=1]
     \bigl(\beta_{+}C(q,u)-\beta_{-}(1-C(q,u))\bigr) \notag\\
 &-\mathbf{1}[j=\mathrm{yes},Y=0]
     \bigl(\bar{\beta}_{+}C(q,u)+\bar{\beta}_{-}(1-C(q,u))\bigr) \notag\\
 &+\mathbf{1}[j=\mathrm{no},Y=0]
     \bigl(\gamma_{+}E(q,u)-\gamma_{-}(1-E(q,u))\bigr) \notag\\
 &-\mathbf{1}[j=\mathrm{no},Y=1]\gamma_{\mathrm{skip}} .
\end{align}
Thus, a ``yes'' judgment must be followed by a correct answer, while a
``no'' judgment must be justified by a valid explanation and is penalized
on answerable examples. The format term is
\begin{equation}
  R_{\mathrm{fmt}}(z)
  =
  \eta_F F(z)-\bar{\eta}_F(1-F(z))
  +
  \eta_S S(z)-\bar{\eta}_S(1-S(z)).
\end{equation}
Table~\ref{tab:fgpo-reward} in Appendix~\ref{app:hyper} lists the implemented values. In implementation, wrong answers and
malformed outputs receive larger penalties than valid ``no'' explanations
receive positive reward, so the model is not rewarded for avoiding
answerable questions by choosing the negative judgment.

\begin{table}[t]
  \centering
  \caption{Overall performance after one TriviaQA knowledge update on Qwen2.5-7B. Train columns use questions seen during the update; Test columns use held-out TriviaQA questions; General reports MMLU and ARC-Easy. $\uparrow$: higher is better; $\downarrow$: lower is better; $\rightarrow 0$: closer to zero is better. Bold and underline indicate the best and second-best results.}
  \label{tab:main}
    \resizebox{\textwidth}{!}{%
\begin{tabular}{@{}lcccccccc@{}}
    \toprule
    & \multicolumn{3}{c}{Train} & \multicolumn{3}{c}{Test} & \multicolumn{2}{c}{General} \\
    \cmidrule(lr){2-4}\cmidrule(lr){5-7}\cmidrule(lr){8-9}
    Method & Acc.$\uparrow$ & AUROC$\uparrow$ & Bias$\rightarrow 0$ & Acc.$\uparrow$ & AUROC$\uparrow$ & Bias$\rightarrow 0$ & MMLU$\uparrow$ & ARC$\uparrow$ \\
    \midrule
    Base                                      & 0.680 & 0.702 & 0.381 & \underline{0.672} & 0.688 & 0.402 & 0.687 & 0.811 \\
    \hspace{1em}+ Prob                        & 0.680 & 0.723 & 0.409 & 0.672 & 0.730 & 0.407 & 0.687 & 0.811 \\
    \hspace{1em}+ N-Prob                      & 0.680 & 0.723 & 0.409 & 0.672 & 0.730 & 0.407 & 0.687 & 0.811 \\
    \hspace{1em}+ EliCal                      & 0.637 & \underline{0.808} & \underline{0.263} & 0.622 & \textbf{0.810} & \textbf{0.254} & 0.709 & 0.504 \\
    \midrule
    Vanilla K-SFT                             & \underline{0.874} & 0.458 & 0.500 & 0.546 & 0.519 & 0.500 & \underline{0.713} & \textbf{0.840} \\
    \hspace{1em}+ Verbal-0                    & 0.874 & 0.491 & 0.543 & 0.546 & 0.478 & 0.551 & 0.713 & 0.840 \\
    \hspace{1em}+ Verbal-10                   & 0.874 & 0.567 & 0.456 & 0.546 & 0.529 & 0.467 & 0.713 & 0.840 \\
    \hspace{1em}+ Prob                        & 0.874 & 0.590 & 0.495 & 0.546 & 0.541 & 0.498 & 0.713 & 0.840 \\
    \hspace{1em}+ N-Prob                      & 0.874 & 0.590 & 0.495 & 0.546 & 0.542 & 0.498 & 0.713 & 0.840 \\
    \hspace{1em}+ Consis-Lex                  & 0.874 & 0.539 & 0.482 & 0.546 & 0.531 & 0.486 & 0.713 & 0.840 \\
    \hspace{1em}+ Consis-Sem                  & 0.874 & 0.553 & 0.472 & 0.546 & 0.567 & 0.467 & 0.713 & 0.840 \\
    \hspace{1em}+ SemEnt                      & 0.874 & 0.538 & 0.481 & 0.546 & 0.526 & 0.504 & 0.713 & 0.840 \\
    \hspace{1em}+ EliCal                      & 0.648 & 0.646 & 0.428 & 0.587 & 0.644 & 0.447 & \textbf{0.723} & 0.688 \\
    \midrule
    Ours                                      & \textbf{0.961} & \textbf{0.817} & \textbf{0.245} & \textbf{0.692} & \underline{0.809} & \underline{0.308} & 0.711 & \underline{0.830} \\
    \bottomrule
  \end{tabular}
  }
\end{table}

\section{Experiments}
\label{sec:exp}

We ask whether FoK can be synchronized after an update (RQ1), which
training stages contribute (RQ2), and whether the result persists across
datasets, sequential updates, and seeds (RQ3). Sections~\ref{subsec:exp-main}, \ref{subsec:exp-rc},
and~\ref{subsec:exp-ablation} address RQ1--RQ2; Section~\ref{subsec:additional-validation} and
Appendix~\ref{app:additional-validation} address RQ3.

\subsection{Evaluation setup}
\label{subsec:exp-settings}

\paragraph{Datasets.}
We use TriviaQA~\citep{joshi-etal-2017-triviaqa} as the main QA dataset, with 5,000 update questions and 1,000 held-out questions. Additional FoK evaluations use SelfAware~\citep{yin2023selfaware}, Natural Questions~\citep{kwiatkowski2019nq}, and WebQuestions~\citep{berant2013webq}. MMLU~\citep{hendryckstest2021} and ARC-Easy~\citep{clark2018think} evaluate general capability.

\paragraph{Baselines.}
We compare direct Yes/No FoK judgment, Vanilla K-SFT, verbalized confidence and 10-bin verbalized confidence~\citep{tian2023}, sequence probability and length-normalized sequence probability, lexical and semantic self-consistency, semantic entropy~\citep{kuhn2023,farquhar2024}, and EliCal~\citep{elical2025}.

\paragraph{Metrics.}
We report accuracy on TriviaQA, MMLU, and ARC-Easy. FoK quality is measured by AUROC~\citep{fawcett2006roc} using the normalized Yes probability, plus two errors thresholded at $J>0.5$: illusion rate (illu., $\Pr(\hat J\!=\!1\mid Y\!=\!0)$) and omission rate (om., $\Pr(\hat J\!=\!0\mid Y\!=\!1)$). We combine them as:
\[
\mathrm{Bias}=
\begin{cases}
\frac{\mathrm{illu.}+\mathrm{om.}}{2}, & \text{if } \mathrm{illu.}\ge \mathrm{om.},\\[2pt]
-\frac{\mathrm{illu.}+\mathrm{om.}}{2}, & \text{if } \mathrm{illu.}< \mathrm{om.}.
\end{cases}
\quad \mathrm{Bias}\in[-1,1].
\]

\paragraph{Implementation details.}
We evaluate on Qwen2.5-7B-Instruct~\citep{qwen25} for the main experiments and on Qwen3-8B~\citep{qwen3} as a thinking-mode backbone. We refer to them as Qwen2.5-7B and Qwen3-8B. We instantiate the three stages of Section~\ref{sec:method} as LoRA adapters on top of the base model. OP-SFT updates the knowledge channel from corrected sampled responses, K$\Delta$-FT trains the FoK channel over the OP-SFT checkpoint trajectory, and FGPO optimizes response generation using both channels. The full data construction, baseline definitions, evaluation protocol, optimizer settings, checkpoint schedule, and inference configuration are listed in Appendix~\ref{app:hyper}.

\subsection{Overall performance}
\label{subsec:exp-main}

\begin{figure}[t]
  \centering
  \includegraphics[width=\linewidth]{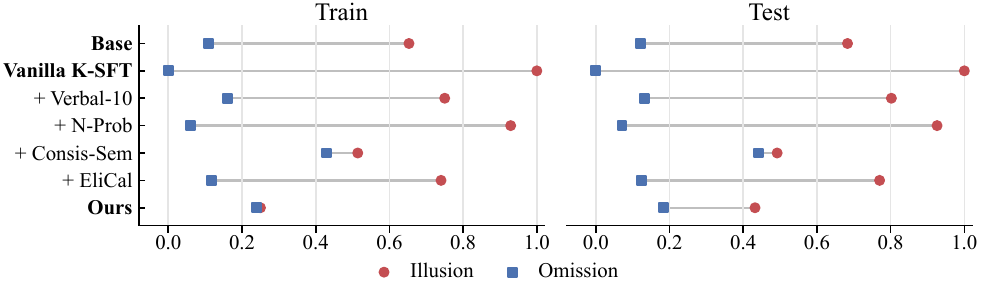}
  \caption{FoK error decomposition for Table~\ref{tab:main}. Red circles show illusion, the conditional rate of false ``can answer'' judgments; blue squares show omission, the conditional rate of false ``cannot answer'' judgments. The horizontal gap shows which error source determines the signed Bias value.}
  \label{fig:fok-error-decomposition}
\end{figure}

\begin{figure}[t]
  \centering
  \includegraphics[width=\linewidth]{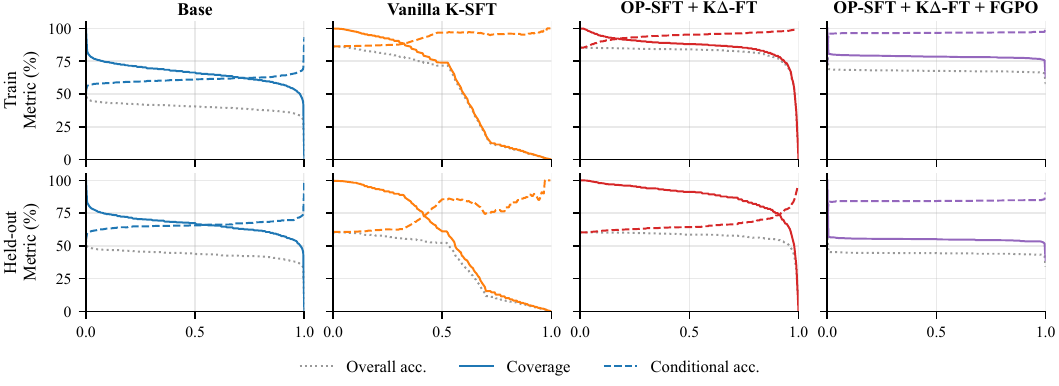}
  \caption{Coverage--accuracy trade-off on Qwen3-8B (thinking mode). Each panel varies the FoK threshold and plots coverage (solid, the fraction of questions the model judges itself able to answer), conditional accuracy on that answered subset (dashed), and overall accuracy with unanswered examples counted as incorrect (dotted).}
  \label{fig:qwen3-selective-kj-grpo}
\end{figure}

We compare our two-channel update with vanilla K-SFT and a suite of existing FoK estimators after a single TriviaQA knowledge update on Qwen2.5-7B. Table~\ref{tab:main} gives the main numbers; Figure~\ref{fig:fok-error-decomposition} decomposes the thresholded errors into illusion and omission. After vanilla K-SFT, in-domain accuracy rises sharply but held-out accuracy falls below the base model, so the update does not simply add knowledge. All existing estimators fall well below their base-model level, and most collapse into one-sided behavior: vanilla K-SFT and confidence-based estimators are dominated by illusion, judging almost every question answerable, while consistency-based estimators trade some illusion for a correspondingly large omission, leaving both error sources substantial. A likely cause is that the knowledge update shifts the output distributions and sampling behavior that these estimators rely on, so a decision boundary calibrated before the update no longer carries the same meaning. This is what Takeaway~2 predicts: under a conventional knowledge update, existing FoK estimators are not a robust way to keep answerability judgments aligned with the updated model.

Our two-channel update shows a different profile. Held-out accuracy now rises above the base model, so the knowledge update generalizes instead of overfitting the update shard. FoK AUROC on the updated model reaches the level of the strongest existing estimator applied to the base model, and Bias is the lowest on the update set and the second lowest on held-out questions, although illusion still exceeds omission on held-out questions (Figure~\ref{fig:fok-error-decomposition}). We attribute this to two design choices. OP-SFT supervises only the corrected answer-relevant span on the model's own sampled responses, which only mildly disturbs FoK: held-out AUROC is 0.603 after OP-SFT alone, compared with 0.519 after vanilla K-SFT (Tables~\ref{tab:main} and~\ref{tab:ablation}); K$\Delta$-FT then trains the FoK channel directly on the changed answerability, instead of reusing a judgment calibrated before the update. Appendix~\ref{app:trajectory} confirms this along the OP-SFT trajectory.

\begin{table}[t]
\centering
\caption{Additional evaluation on Qwen2.5-7B. Left: all methods start from the same OP-SFT knowledge checkpoint; K$\Delta$-FT adds the FoK adapter, so its combined answering state differs. Right: $\varphi_1$ is frozen after round 1; $\varphi_i$ is retrained over each round's knowledge trajectory. Each round adds a disjoint shard from 5,000 update questions; all test columns use the same 1,000 held-out questions.}
\label{tab:additional-main}
\begin{subtable}[t]{0.49\linewidth}
\centering
\caption{FoK estimators after OP-SFT.}
\label{tab:opsft-controlled}
\begin{tabular}{@{}lcc@{}}
\toprule
Estimator & Test Acc. & AUROC\\
\midrule
Direct judgment & 0.679 & 0.603\\
Prob & 0.679 & 0.571\\
EliCal & 0.679 & 0.618\\
K$\Delta$-FT & 0.699 & 0.770\\
\bottomrule
\end{tabular}
\end{subtable}\hfill
\begin{subtable}[t]{0.49\linewidth}
\centering
\caption{Three sequential TriviaQA updates.}
\label{tab:sequential-main}
\begin{tabular}{@{}lccc@{}}
\toprule
Round & Test Acc. & $\varphi_1$ AUC & $\varphi_i$ AUC\\
\midrule
1 & 0.659 & 0.798 & 0.798\\
2 & 0.644 & 0.776 & 0.800\\
3 & 0.661 & 0.754 & 0.797\\
\bottomrule
\end{tabular}
\end{subtable}
\end{table}

\begin{table}[t]
  \caption{Held-out stage-wise ablation on Qwen2.5-7B.
  OP-SFT updates knowledge, K$\Delta$-FT adapts FoK, and FGPO optimizes
  judgment-guided generation. Higher accuracy and AUROC and lower error
  rates are better; Bias closer to zero is better. Update-set results
  are in Table~\ref{tab:ablation-update-set}.}
  \label{tab:ablation}
  \centering
  \begin{tabular}{@{}lccccc@{}}
    \toprule
    Stage & Acc.$\uparrow$ & AUROC$\uparrow$ & Illu.$\downarrow$ & Om.$\downarrow$ & Bias$\to0$\\
    \midrule
    Base & 0.672 & 0.688 & 0.683 & 0.122 & 0.402 \\
    Stage 1: + OP-SFT & 0.679 & 0.603 & 0.601 & 0.283 & 0.442 \\
    Stage 2: + K$\Delta$-FT & 0.699 & 0.770 & 0.478 & 0.187 & 0.333 \\
    Stage 3: + FGPO & 0.692 & 0.809 & 0.432 & 0.184 & 0.308 \\
    \bottomrule
  \end{tabular}
\end{table}

\subsection{Coverage-accuracy trade-off under FoK judgment}
\label{subsec:exp-rc}

Table~\ref{tab:main} shows that FoK quality is recoverable; it remains to test whether the recovered FoK actually drives the decision of when to answer. We therefore vary the FoK threshold on Qwen3-8B and plot the coverage-accuracy curves in Figure~\ref{fig:qwen3-selective-kj-grpo}. For Base and Vanilla K-SFT the curves are nearly flat on both splits: lowering coverage does not raise accuracy, so their FoK scores do not rank answerability. OP-SFT + K$\Delta$-FT restores a monotonic trade-off, confirming that the learned FoK channel is consistent with answerability after the update. FGPO flattens the curves again, but for a different reason: the scores collapse to $0$ or $1$, and the threshold becomes irrelevant once the policy has committed to answer or abstain. K$\Delta$-FT thus provides a tunable coverage--accuracy trade-off, while FGPO answers more selectively, trading coverage for accuracy on answered questions.

\subsection{Controlled comparison and sequential updates}
\label{subsec:additional-validation}

\paragraph{Accuracy-preserving update.}
Table~\ref{tab:opsft-controlled} tests whether avoiding destructive K-SFT
alone suffices. After OP-SFT, direct judgment, Prob, and EliCal reach
AUROC 0.603, 0.571, and 0.618, respectively; K$\Delta$-FT reaches 0.770.
Thus, FoK adaptation helps even when held-out accuracy is preserved.
When no update is applied, strong unadapted estimators remain competitive and the accuracy gain of the update itself is modest: Base + EliCal reaches 0.810 AUROC in Table~\ref{tab:main}, compared with 0.809 for the full method. Once an update occurs, the unadapted estimators desynchronize from the new knowledge state, and the continually trained FoK channel is what preserves judgment quality.

\paragraph{Sequential updates and broader evaluation.}
Across three updates, the per-round FoK channel stays near 0.80 AUROC
while the frozen channel falls to 0.754
(Table~\ref{tab:sequential-main}). Cross-dataset tests on SelfAware, Natural Questions, and
WebQuestions, plus a Natural Questions replication, extend the
factual-QA evaluation. Independent Stage-1+2 runs yield test AUROC
$0.777\pm0.009$ (four seeds); FGPO yields $0.811\pm0.003$ (three seeds
from one shared initialization). Values are means and sample SDs. Appendix~\ref{app:additional-validation} reports all runs,
cross-dataset results, and a MEMIT desynchronization test.

\subsection{Ablation study}
\label{subsec:exp-ablation}

To locate where the FoK improvement in Table~\ref{tab:main} comes from, we ablate the three stages cumulatively: OP-SFT alone, then OP-SFT + K$\Delta$-FT, then the full pipeline with FGPO (Table~\ref{tab:ablation}). OP-SFT alone raises update-set accuracy from 0.680 to 0.966 (Table~\ref{tab:ablation-update-set}) but leaves the FoK judgment misaligned with the updated knowledge state. Adding K$\Delta$-FT is the main source of the FoK gain: AUROC rises and Bias drops while held-out accuracy changes little (0.679 to 0.699), supporting the design choice of training a dedicated FoK channel over knowledge-delta states rather than relying on knowledge learning alone. FGPO raises held-out AUROC from 0.770 to 0.809 and incorporates the judgment into generation. On the separate Qwen3-8B evaluation, its nearly binary judgments explain the flattened threshold curves in Figure~\ref{fig:qwen3-selective-kj-grpo}.

\section{Related work}
\label{sec:related}

\paragraph{Functional metacognition.}
Metacognitive sensitivity and calibration are distinct quantities
\citep{fleming2014metacognition}. Human--LLM comparisons therefore
require explicit behavioral definitions rather than equating similar
reports with identical cognitive processes
\citep{steyvers2026metacognition}. Recent work studies competing
confidence biases under external advice \citep{kumaran2026biases} and
how verbal confidence is computed from answer-related representations
\citep{kumaran2026verbal}. Our focus is correctness-supervised
answerability judgment under changing model parameters. Koriat's
cue-accessibility account and the cue-familiarity account of FoK
\citep{koriat1993,reder1992,koriat2001} motivate our context analyses.

\paragraph{LLM uncertainty and hallucination.}
LLMs often generate fluent but factually false outputs, a failure studied
as hallucination and factuality \citep{ji2023survey,huang2023survey}.
Theoretical analyses show that this failure cannot be removed by
pretraining alone: rare facts impose a Good--Turing-style lower bound on
calibrated error, and common training and evaluation procedures can
reward guessing rather than explicit uncertainty \citep{kalai2023,
kalai2025}. This motivates uncertainty estimation and answerability
judgment in LLMs \citep{geng2023}. Existing methods estimate uncertainty
from generated outputs, including $P(\text{True})$ and $P(\text{IK})$
(probability that the model ``knows'' the answer),
verbalized confidence, explicit ``I don't know'' (IDK) training,
semantic entropy, and
self-consistency \citep{kadavath2022,lin2022,tian2023,xiong2024,
cheng2024idk,kuhn2023,farquhar2024,wang2023,kossen2024}. Other methods
read uncertainty from internal representations, including SAPLMA, CCS,
INSIDE, MIND, C$^3$, and internal-state probes
\citep{azaria2023,burns2022,chen2024inside,su2024,c3_2025,
slobodkin2023}. A related line trains answerability behavior directly:
R-Tuning teaches models to state that they do not know
\citep{zhang2023rtuning}; US-Tuning, CRaFT, and GRAIT analyze conflicts
between answer learning and negative answerability supervision
\citep{liang2024know,wang2024craft,chen2025grait}; EliCal uses
annotation-efficient elicitation and calibration for honesty
\citep{elical2025}. Inference-time factuality methods such as ITI and
DoLa intervene without additional training \citep{li2023iti,chuang2023},
while TruthfulQA, HaluEval, FActScore, and SimpleQA evaluate factual
support in generated answers \citep{lin2021,li2023halueval,min2023,
openai2024}. These works primarily ask whether a given model can express
or reveal uncertainty about its current answers. We instead ask whether
the FoK judgment tracks post-training updates to the model's knowledge.

\paragraph{Knowledge editing and continual learning.}
Post-training updates, knowledge editing, and continual learning change
what an LLM knows after pretraining. ROME and MEMIT edit factual
associations and evaluate edit success, locality, portability, and
retention \citep{meng2022rome,meng2023memit,fu2025mmke}. Continual-learning
methods, including regularization and replay-based approaches, aim to
preserve answer accuracy across sequential updates
\citep{kirkpatrick2017,jang2022,luo2023}. These methods focus on the
update operator and on answer correctness, and usually do not test
whether answerability judgment changes with the updated knowledge.
We instead treat FoK as a training target co-learned with each update,
so the answerability judgment moves with the model rather than lagging
behind on a fixed checkpoint.

\section{Conclusion}
\label{sec:conclusion}

We studied how to keep functional FoK aligned with an LLM's answerability after knowledge updates. In factual QA, knowledge updating alone does not preserve this alignment. Our two-channel framework trains knowledge and FoK separately and then optimizes the judgment together with the subsequent response. Controlled OP-SFT comparisons, cross-dataset evaluation, and three sequential updates support the usefulness of explicit FoK adaptation, with seed variation quantifying its stability in the main setting.

\begin{ack}
This research was partially supported by National Natural Science Foundation of China (No.62502404), Hong Kong Research Grants Council (Research Impact Fund No.R1015-23, Collaborative Research Fund No.C1043-24GF, RGC Research Fellow Scheme No.RFS2627-1S03, General Research Fund No.11218325, No.11212926), Institute of Digital Medicine of City University of Hong Kong (No.9229503), and Tencent (CCF-Tencent Open Fund, Tencent Rhino-Bird Focused Research Program, Tencent University Cooperation Project).

\end{ack}

\bibliographystyle{plainnat}
\bibliography{custom}

\begin{thebibliography}{78}
\providecommand{\natexlab}[1]{#1}
\providecommand{\url}[1]{\texttt{#1}}
\expandafter\ifx\csname urlstyle\endcsname\relax
  \providecommand{\doi}[1]{doi: #1}\else
  \providecommand{\doi}{doi: \begingroup \urlstyle{rm}\Url}\fi

\bibitem[Azaria and Mitchell(2023)]{azaria2023}
Amos Azaria and Tom Mitchell.
\newblock The internal state of an {LLM} knows when it's lying.
\newblock \emph{arXiv preprint arXiv:2304.13734}, 2023.

\bibitem[Berant et~al.(2013)Berant, Chou, Frostig, and Liang]{berant2013webq}
Jonathan Berant, Andrew Chou, Roy Frostig, and Percy Liang.
\newblock Semantic parsing on {Freebase} from question-answer pairs.
\newblock In \emph{Proceedings of the 2013 Conference on Empirical Methods in
  Natural Language Processing}, pages 1533--1544, 2013.
\newblock URL \url{https://aclanthology.org/D13-1160/}.

\bibitem[Burns et~al.(2023)Burns, Ye, Klein, and Steinhardt]{burns2022}
Collin Burns, Haotian Ye, Dan Klein, and Jacob Steinhardt.
\newblock Discovering latent knowledge in language models without supervision.
\newblock In \emph{International Conference on Learning Representations
  (ICLR)}, 2023.

\bibitem[Chen et~al.(2024)Chen, Liu, Chen, Gu, Wu, Tao, Fu, and
  Ye]{chen2024inside}
Chao Chen, Kai Liu, Ze~Chen, Yi~Gu, Yue Wu, Mingyuan Tao, Zhihang Fu, and
  Jieping Ye.
\newblock {INSIDE}: {LLM}s' internal states retain the power of hallucination
  detection.
\newblock In \emph{International Conference on Learning Representations}, 2024.
\newblock URL \url{https://openreview.net/forum?id=Zj12nzlQbz}.

\bibitem[Chuang et~al.(2024)Chuang, Xie, Luo, Kim, Glass, and He]{chuang2023}
Yung-Sung Chuang, Yujia Xie, Hongyin Luo, Yoon Kim, James~R. Glass, and
  Pengcheng He.
\newblock {DoLa}: Decoding by contrasting layers improves factuality in large
  language models.
\newblock In \emph{International Conference on Learning Representations}, 2024.
\newblock URL \url{https://openreview.net/forum?id=Th6NyL07na}.

\bibitem[Clark et~al.(2018)Clark, Cowhey, Etzioni, Khot, Sabharwal, Schoenick,
  and Tafjord]{clark2018think}
Peter Clark, Isaac Cowhey, Oren Etzioni, Tushar Khot, Ashish Sabharwal, Carissa
  Schoenick, and Oyvind Tafjord.
\newblock Think you have solved question answering? try {ARC}, the {AI2}
  reasoning challenge, 2018.

\bibitem[Cobbe et~al.(2021)Cobbe, Kosaraju, Bavarian, Hilton, Nakano, Hesse,
  and Schulman]{cobbe2021training}
Karl Cobbe, Vineet Kosaraju, Mohammad Bavarian, Jacob Hilton, Reiichiro Nakano,
  Christopher Hesse, and John Schulman.
\newblock Training verifiers to solve math word problems, 2021.

\bibitem[Cohen et~al.(2024)Cohen, Dobler, Biran, and de~Melo]{cheng2024idk}
Roi Cohen, Konstantin Dobler, Eden Biran, and Gerard de~Melo.
\newblock I don't know: Explicit modeling of uncertainty with an {[IDK]} token.
\newblock In \emph{Advances in Neural Information Processing Systems}, 2024.
\newblock \doi{10.52202/079017-0349}.

\bibitem[Farquhar et~al.(2024)Farquhar, Kossen, Kuhn, and Gal]{farquhar2024}
Sebastian Farquhar, Jannik Kossen, Lorenz Kuhn, and Yarin Gal.
\newblock Detecting hallucinations in large language models using semantic
  entropy.
\newblock \emph{Nature}, 630:\penalty0 625--630, 2024.
\newblock \doi{10.1038/s41586-024-07421-0}.

\bibitem[Fawcett(2006)]{fawcett2006roc}
Tom Fawcett.
\newblock An introduction to {ROC} analysis.
\newblock \emph{Pattern Recognition Letters}, 27\penalty0 (8):\penalty0
  861--874, 2006.
\newblock \doi{10.1016/j.patrec.2005.10.010}.

\bibitem[Fleming and Lau(2014)]{fleming2014metacognition}
Stephen~M. Fleming and Hakwan~C. Lau.
\newblock How to measure metacognition.
\newblock \emph{Frontiers in Human Neuroscience}, 8:\penalty0 443, 2014.
\newblock \doi{10.3389/fnhum.2014.00443}.

\bibitem[Fu et~al.(2025{\natexlab{a}})Fu, Li, Wu, Wang, Dong, Zhao, Zhao, Guo,
  and Tang]{fu2025unictr}
Zichuan Fu, Xiangyang Li, Chuhan Wu, Yichao Wang, Kuicai Dong, Xiangyu Zhao,
  Mengchen Zhao, Huifeng Guo, and Ruiming Tang.
\newblock A unified framework for multi-domain {CTR} prediction via large
  language models.
\newblock \emph{ACM Transactions on Information Systems}, 43\penalty0
  (5):\penalty0 117, 2025{\natexlab{a}}.

\bibitem[Fu et~al.(2025{\natexlab{b}})Fu, Wu, Li, Zhang, Zheng, Ming, Wang,
  Wang, and Zhao]{fu2025mmke}
Zichuan Fu, Xian Wu, Guojing Li, Yingying Zhang, Yefeng Zheng, Tianshi Ming,
  Yejing Wang, Wanyu Wang, and Xiangyu Zhao.
\newblock Model merging for knowledge editing.
\newblock In \emph{Proceedings of the 63rd Annual Meeting of the Association
  for Computational Linguistics (Volume 6: Industry Track)},
  2025{\natexlab{b}}.

\bibitem[Fu et~al.(2025{\natexlab{c}})Fu, Wu, Wang, Wang, Ye, Yin, Chang,
  Zheng, and Zhao]{fu2025tfmerge}
Zichuan Fu, Xian Wu, Yejing Wang, Wanyu Wang, Shanshan Ye, Hongzhi Yin,
  Yi~Chang, Yefeng Zheng, and Xiangyu Zhao.
\newblock Training-free {LLM} merging for multi-task learning.
\newblock In \emph{Proceedings of the 63rd Annual Meeting of the Association
  for Computational Linguistics (Volume 1: Long Papers)}, 2025{\natexlab{c}}.

\bibitem[Fu et~al.(2026)Fu, Wu, Li, Wang, Chen, Zhao, Luo, Yan, Zheng, and
  Zhao]{fu2026tandem}
Zichuan Fu, Xian Wu, Guojing Li, Yejing Wang, Yijun Chen, Zihao Zhao, Yixuan
  Luo, Hanyu Yan, Yefeng Zheng, and Xiangyu Zhao.
\newblock Tandem: Riding together with large and small language models for
  efficient reasoning.
\newblock In \emph{Findings of the Association for Computational Linguistics:
  ACL 2026}, pages 42286--42302, 2026.

\bibitem[Geng et~al.(2024)Geng, Cai, Wang, Koeppl, Nakov, and
  Gurevych]{geng2023}
Jiahui Geng, Fengyu Cai, Yuxia Wang, Heinz Koeppl, Preslav Nakov, and Iryna
  Gurevych.
\newblock A survey of confidence estimation and calibration in large language
  models.
\newblock In \emph{Proceedings of the 2024 Conference of the North American
  Chapter of the Association for Computational Linguistics: Human Language
  Technologies (Volume 1: Long Papers)}, pages 6577--6595. Association for
  Computational Linguistics, 2024.
\newblock \doi{10.18653/v1/2024.naacl-long.366}.
\newblock URL \url{https://aclanthology.org/2024.naacl-long.366/}.

\bibitem[Hart(1965)]{hart1965}
Joseph~T. Hart.
\newblock Memory and the feeling-of-knowing experience.
\newblock \emph{Journal of Educational Psychology}, 56\penalty0 (4):\penalty0
  208--216, 1965.

\bibitem[Hendrycks et~al.(2021{\natexlab{a}})Hendrycks, Burns, Basart, Zou,
  Mazeika, Song, and Steinhardt]{hendryckstest2021}
Dan Hendrycks, Collin Burns, Steven Basart, Andy Zou, Mantas Mazeika, Dawn
  Song, and Jacob Steinhardt.
\newblock Measuring massive multitask language understanding.
\newblock In \emph{International Conference on Learning Representations},
  2021{\natexlab{a}}.

\bibitem[Hendrycks et~al.(2021{\natexlab{b}})Hendrycks, Burns, Kadavath, Arora,
  Basart, Tang, Song, and Steinhardt]{hendrycks2021math}
Dan Hendrycks, Collin Burns, Saurav Kadavath, Akul Arora, Steven Basart, Eric
  Tang, Dawn Song, and Jacob Steinhardt.
\newblock Measuring mathematical problem solving with the {MATH} dataset.
\newblock In \emph{Proceedings of the Neural Information Processing Systems
  Track on Datasets and Benchmarks}, 2021{\natexlab{b}}.

\bibitem[Huang et~al.(2024)Huang, Yu, Ma, Zhong, Feng, Wang, Chen, Peng, Feng,
  Qin, and Liu]{huang2023survey}
Lei Huang, Weijiang Yu, Weitao Ma, Weihong Zhong, Zhangyin Feng, Haotian Wang,
  Qianglong Chen, Weihua Peng, Xiaocheng Feng, Bing Qin, and Ting Liu.
\newblock A survey on hallucination in large language models: Principles,
  taxonomy, challenges, and open questions.
\newblock \emph{ACM Transactions on Information Systems}, 2024.
\newblock \doi{10.1145/3703155}.

\bibitem[Jang et~al.(2022)Jang, Ye, Yang, Shin, Han, Kim, Choi, and
  Seo]{jang2022}
Joel Jang, Seonghyeon Ye, Sohee Yang, Joongbo Shin, Janghoon Han, Gyeonghun
  Kim, Stanley~Jungkyu Choi, and Minjoon Seo.
\newblock Towards continual knowledge learning of language models.
\newblock In \emph{International Conference on Learning Representations
  (ICLR)}, 2022.

\bibitem[Jeong et~al.(2024)Jeong, Baek, Cho, Hwang, and Park]{yang2024}
Soyeong Jeong, Jinheon Baek, Sukmin Cho, Sung~Ju Hwang, and Jong~C. Park.
\newblock Adaptive-{RAG}: Learning to adapt retrieval-augmented large language
  models through question complexity.
\newblock In \emph{Proceedings of the 2024 Conference of the North American
  Chapter of the Association for Computational Linguistics: Human Language
  Technologies (Volume 1: Long Papers)}, pages 7036--7050. Association for
  Computational Linguistics, 2024.
\newblock \doi{10.18653/v1/2024.naacl-long.389}.
\newblock URL \url{https://aclanthology.org/2024.naacl-long.389/}.

\bibitem[Ji et~al.(2023)Ji, Lee, Frieske, Yu, Su, Xu, Ishii, Bang, Madotto, and
  Fung]{ji2023survey}
Ziwei Ji, Nayeon Lee, Rita Frieske, Tiezheng Yu, Dan Su, Yan Xu, Etsuko Ishii,
  Yejin Bang, Andrea Madotto, and Pascale Fung.
\newblock Survey of hallucination in natural language generation.
\newblock \emph{ACM Computing Surveys}, 55\penalty0 (12):\penalty0 1--38, 2023.
\newblock \doi{10.1145/3571730}.

\bibitem[Jia et~al.(2024)Jia, Liu, Zhao, Li, Hao, Wang, and Yin]{jia2024mill}
Pengyue Jia, Yiding Liu, Xiangyu Zhao, Xiaopeng Li, Changying Hao, Shuaiqiang
  Wang, and Dawei Yin.
\newblock {MILL}: Mutual verification with large language models for zero-shot
  query expansion.
\newblock In \emph{Proceedings of the 2024 Conference of the North American
  Chapter of the Association for Computational Linguistics: Human Language
  Technologies (Volume 1: Long Papers)}, pages 2498--2518, 2024.

\bibitem[Joshi et~al.(2017)Joshi, Choi, Weld, and
  Zettlemoyer]{joshi-etal-2017-triviaqa}
Mandar Joshi, Eunsol Choi, Daniel Weld, and Luke Zettlemoyer.
\newblock {T}rivia{QA}: A large scale distantly supervised challenge dataset
  for reading comprehension.
\newblock In Regina Barzilay and Min-Yen Kan, editors, \emph{Proceedings of the
  55th Annual Meeting of the Association for Computational Linguistics (Volume
  1: Long Papers)}, pages 1601--1611, Vancouver, Canada, July 2017. Association
  for Computational Linguistics.
\newblock \doi{10.18653/v1/P17-1147}.
\newblock URL \url{https://aclanthology.org/P17-1147/}.

\bibitem[Kadavath et~al.(2022)Kadavath, Conerly, Askell, Henighan, Drain,
  Perez, Schiefer, Hatfield-Dodds, DasSarma, Tran-Johnson, Johnston, El-Showk,
  Jones, Elhage, Hume, Chen, Bai, Bowman, Fort, Ganguli, Hernandez, Jacobson,
  Kernion, Kravec, Lovitt, Ndousse, Olsson, Ringer, Amodei, Brown, Clark,
  Joseph, Mann, McCandlish, Olah, and Kaplan]{kadavath2022}
Saurav Kadavath, Tom Conerly, Amanda Askell, Tom Henighan, Dawn Drain, Ethan
  Perez, Nicholas Schiefer, Zac Hatfield-Dodds, Nova DasSarma, Eli
  Tran-Johnson, Scott Johnston, Sheer El-Showk, Andy Jones, Nelson Elhage,
  Tristan Hume, Anna Chen, Yuntao Bai, Sam Bowman, Stanislav Fort, Deep
  Ganguli, Danny Hernandez, Josh Jacobson, Jackson Kernion, Shauna Kravec,
  Liane Lovitt, Kamal Ndousse, Catherine Olsson, Sam Ringer, Dario Amodei, Tom
  Brown, Jack Clark, Nicholas Joseph, Ben Mann, Sam McCandlish, Chris Olah, and
  Jared Kaplan.
\newblock Language models (mostly) know what they know.
\newblock \emph{arXiv preprint arXiv:2207.05221}, 2022.
\newblock \doi{10.48550/arXiv.2207.05221}.

\bibitem[Kalai and Vempala(2024)]{kalai2023}
Adam~Tauman Kalai and Santosh~S. Vempala.
\newblock Calibrated language models must hallucinate.
\newblock In \emph{Proceedings of the 56th Annual ACM Symposium on Theory of
  Computing}, 2024.
\newblock \doi{10.1145/3618260.3649777}.

\bibitem[Kalai et~al.(2025)Kalai, Nachum, Vempala, and Zhang]{kalai2025}
Adam~Tauman Kalai, Ofir Nachum, Santosh~S. Vempala, and Edwin Zhang.
\newblock Why language models hallucinate.
\newblock \emph{arXiv preprint arXiv:2509.04664}, 2025.
\newblock \doi{10.48550/arXiv.2509.04664}.

\bibitem[Kirkpatrick et~al.(2017)Kirkpatrick, Pascanu, Rabinowitz, Veness,
  Desjardins, Rusu, Milan, Quan, Ramalho, Grabska-Barwinska, Hassabis, Clopath,
  Kumaran, and Hadsell]{kirkpatrick2017}
James Kirkpatrick, Razvan Pascanu, Neil Rabinowitz, Joel Veness, Guillaume
  Desjardins, Andrei~A. Rusu, Kieran Milan, John Quan, Tiago Ramalho, Agnieszka
  Grabska-Barwinska, Demis Hassabis, Claudia Clopath, Dharshan Kumaran, and
  Raia Hadsell.
\newblock Overcoming catastrophic forgetting in neural networks.
\newblock \emph{Proceedings of the National Academy of Sciences}, 114\penalty0
  (13):\penalty0 3521--3526, 2017.
\newblock \doi{10.1073/pnas.1611835114}.

\bibitem[Koriat(1993)]{koriat1993}
Asher Koriat.
\newblock How do we know that we know? {T}he accessibility model of the feeling
  of knowing.
\newblock \emph{Psychological Review}, 100\penalty0 (4):\penalty0 609--639,
  1993.

\bibitem[Koriat and Levy-Sadot(2001)]{koriat2001}
Asher Koriat and Ravit Levy-Sadot.
\newblock The combined contributions of the cue-familiarity and accessibility
  heuristics to feelings of knowing.
\newblock \emph{Journal of Experimental Psychology: Learning, Memory, and
  Cognition}, 27\penalty0 (1):\penalty0 34--53, 2001.

\bibitem[Kossen et~al.(2024)Kossen, Han, Razzak, Schut, Malik, and
  Gal]{kossen2024}
Jannik Kossen, Jiatong Han, Muhammed Razzak, Lisa Schut, Shreshth~A. Malik, and
  Yarin Gal.
\newblock Semantic entropy probes: Robust and cheap hallucination detection in
  {LLM}s.
\newblock \emph{arXiv preprint arXiv:2406.15927}, 2024.
\newblock \doi{10.48550/arXiv.2406.15927}.

\bibitem[Kuhn et~al.(2023)Kuhn, Gal, and Farquhar]{kuhn2023}
Lorenz Kuhn, Yarin Gal, and Sebastian Farquhar.
\newblock Semantic uncertainty: Linguistic invariances for uncertainty
  estimation in natural language generation.
\newblock In \emph{International Conference on Learning Representations}, 2023.
\newblock URL \url{https://openreview.net/forum?id=VD-AYtP0dve}.

\bibitem[Kumaran et~al.(2026{\natexlab{a}})Kumaran, Conmy, Barbero, Osindero,
  Patraucean, and Veli{\v{c}}kovi{\'c}]{kumaran2026verbal}
Dharshan Kumaran, Arthur Conmy, Federico Barbero, Simon Osindero, Viorica
  Patraucean, and Petar Veli{\v{c}}kovi{\'c}.
\newblock How do {LLMs} compute verbal confidence, 2026{\natexlab{a}}.
\newblock URL \url{https://arxiv.org/abs/2603.17839}.

\bibitem[Kumaran et~al.(2026{\natexlab{b}})Kumaran, Fleming, Markeeva, Heyward,
  Banino, Mathur, Pascanu, Osindero, De~Martino, Veli{\v{c}}kovi{\'c}, and
  Patraucean]{kumaran2026biases}
Dharshan Kumaran, Stephen~M. Fleming, Larisa Markeeva, Joe Heyward, Andrea
  Banino, Mrinal Mathur, Razvan Pascanu, Simon Osindero, Benedetto De~Martino,
  Petar Veli{\v{c}}kovi{\'c}, and Viorica Patraucean.
\newblock Competing biases underlie overconfidence and underconfidence in
  {LLMs}.
\newblock \emph{Nature Machine Intelligence}, 8:\penalty0 614--627,
  2026{\natexlab{b}}.
\newblock \doi{10.1038/s42256-026-01217-9}.

\bibitem[Kwiatkowski et~al.(2019)Kwiatkowski, Palomaki, Redfield, Collins,
  Parikh, Alberti, Epstein, Polosukhin, Devlin, Lee, Toutanova, Jones, Kelcey,
  Chang, Dai, Uszkoreit, Le, and Petrov]{kwiatkowski2019nq}
Tom Kwiatkowski, Jennimaria Palomaki, Olivia Redfield, Michael Collins, Ankur
  Parikh, Chris Alberti, Danielle Epstein, Illia Polosukhin, Jacob Devlin,
  Kenton Lee, Kristina Toutanova, Llion Jones, Matthew Kelcey, Ming-Wei Chang,
  Andrew~M. Dai, Jakob Uszkoreit, Quoc Le, and Slav Petrov.
\newblock Natural questions: A benchmark for question answering research.
\newblock \emph{Transactions of the Association for Computational Linguistics},
  7:\penalty0 452--466, 2019.
\newblock \doi{10.1162/tacl_a_00276}.
\newblock URL \url{https://aclanthology.org/Q19-1026/}.

\bibitem[Levy et~al.(2017)Levy, Seo, Choi, and Zettlemoyer]{levy2017zsre}
Omer Levy, Minjoon Seo, Eunsol Choi, and Luke Zettlemoyer.
\newblock Zero-shot relation extraction via reading comprehension.
\newblock In \emph{Proceedings of the 21st Conference on Computational Natural
  Language Learning (CoNLL 2017)}, pages 333--342, 2017.

\bibitem[Li et~al.(2025{\natexlab{a}})Li, Tang, and Yang]{liang2024know}
Jiaqi Li, Yixuan Tang, and Yi~Yang.
\newblock Know the unknown: An uncertainty-sensitive method for {LLM}
  instruction tuning.
\newblock In \emph{Findings of the Association for Computational Linguistics:
  ACL 2025}, pages 2972--2989. Association for Computational Linguistics,
  2025{\natexlab{a}}.
\newblock \doi{10.18653/v1/2025.findings-acl.153}.
\newblock URL \url{https://aclanthology.org/2025.findings-acl.153/}.

\bibitem[Li et~al.(2023{\natexlab{a}})Li, Cheng, Zhao, Nie, and
  Wen]{li2023halueval}
Junyi Li, Xiaoxue Cheng, Xin Zhao, Jian-Yun Nie, and Ji-Rong Wen.
\newblock {HaluEval}: A large-scale hallucination evaluation benchmark for
  large language models.
\newblock In \emph{Proceedings of the 2023 Conference on Empirical Methods in
  Natural Language Processing}, pages 6449--6464. Association for Computational
  Linguistics, 2023{\natexlab{a}}.
\newblock \doi{10.18653/v1/2023.emnlp-main.397}.
\newblock URL \url{https://aclanthology.org/2023.emnlp-main.397/}.

\bibitem[Li et~al.(2023{\natexlab{b}})Li, Patel, Vi{\'e}gas, Pfister, and
  Wattenberg]{li2023iti}
Kenneth Li, Oam Patel, Fernanda Vi{\'e}gas, Hanspeter Pfister, and Martin
  Wattenberg.
\newblock Inference-time intervention: Eliciting truthful answers from a
  language model.
\newblock In \emph{Advances in Neural Information Processing Systems},
  2023{\natexlab{b}}.

\bibitem[Li et~al.(2025{\natexlab{b}})Li, Jia, Xu, Wen, Zhang, Zhang, Wang,
  Wang, Du, Li, Liu, Guo, Tang, and Zhao]{li2025personalization}
Xiaopeng Li, Pengyue Jia, Derong Xu, Yi~Wen, Yingyi Zhang, Wenlin Zhang, Wanyu
  Wang, Yichao Wang, Zhaocheng Du, Xiangyang Li, Yong Liu, Huifeng Guo, Ruiming
  Tang, and Xiangyu Zhao.
\newblock A survey of personalization: From {RAG} to agent.
\newblock \emph{arXiv preprint arXiv:2504.10147}, 2025{\natexlab{b}}.
\newblock \doi{10.48550/arXiv.2504.10147}.

\bibitem[Li et~al.(2025{\natexlab{c}})Li, Cai, Kong, Chen, Chen, Yang, Zhang,
  Li, Wu, Chen, Qu, Ye, Su, Ma, Yan, Xia, Shi, Wang, Zhao, Zhao, Xiong, Wang,
  and Yin]{li2025aisearch}
Yuchen Li, Hengyi Cai, Rui Kong, Xinran Chen, Jiamin Chen, Jun Yang, Haojie
  Zhang, Jiayi Li, Jiayi Wu, Yiqun Chen, Changle Qu, Wenwen Ye, Lixin Su, Xinyu
  Ma, Lingyong Yan, Long Xia, Daiting Shi, Junfeng Wang, Xiangyu Zhao, Jiashu
  Zhao, Haoyi Xiong, Shuaiqiang Wang, and Dawei Yin.
\newblock Towards {AI} search paradigm.
\newblock \emph{arXiv preprint arXiv:2506.17188}, 2025{\natexlab{c}}.
\newblock \doi{10.48550/arXiv.2506.17188}.

\bibitem[Lin et~al.(2022{\natexlab{a}})Lin, Hilton, and Evans]{lin2021}
Stephanie Lin, Jacob Hilton, and Owain Evans.
\newblock {TruthfulQA}: Measuring how models mimic human falsehoods.
\newblock In \emph{Proceedings of the 60th Annual Meeting of the Association
  for Computational Linguistics (Volume 1: Long Papers)}, pages 3214--3252.
  Association for Computational Linguistics, 2022{\natexlab{a}}.
\newblock \doi{10.18653/v1/2022.acl-long.229}.
\newblock URL \url{https://aclanthology.org/2022.acl-long.229/}.

\bibitem[Lin et~al.(2022{\natexlab{b}})Lin, Hilton, and Evans]{lin2022}
Stephanie Lin, Jacob Hilton, and Owain Evans.
\newblock Teaching models to express their uncertainty in words.
\newblock \emph{Transactions on Machine Learning Research}, 2022{\natexlab{b}}.

\bibitem[Liu et~al.(2024)Liu, Wu, Zhao, Zhu, Zhang, Tian, and
  Zheng]{liu2024leader}
Qidong Liu, Xian Wu, Xiangyu Zhao, Yuanshao Zhu, Zijian Zhang, Feng Tian, and
  Yefeng Zheng.
\newblock Large language model distilling medication recommendation model.
\newblock \emph{arXiv preprint arXiv:2402.02803}, 2024.
\newblock \doi{10.48550/arXiv.2402.02803}.

\bibitem[Liu et~al.(2025)Liu, Zhao, Wang, Zhang, Zhong, Chen, Li, Huang, and
  Tian]{liu2025llm4cdsr}
Qidong Liu, Xiangyu Zhao, Yejing Wang, Zijian Zhang, Howard Zhong, Chong Chen,
  Xiang Li, Wei Huang, and Feng Tian.
\newblock Bridge the domains: Large language models enhanced cross-domain
  sequential recommendation.
\newblock In \emph{Proceedings of the 48th International ACM SIGIR Conference
  on Research and Development in Information Retrieval}, 2025.
\newblock \doi{10.1145/3726302.3729911}.

\bibitem[Liu et~al.(2023)Liu, Cai, Sun, Wang, Jiang, Zheng, Gai, Jiang, Zhao,
  and Zhang]{liu2023latent}
Shuchang Liu, Qingpeng Cai, Bowen Sun, Yuhao Wang, Ji~Jiang, Dong Zheng, Kun
  Gai, Peng Jiang, Xiangyu Zhao, and Yongfeng Zhang.
\newblock Exploration and regularization of the latent action space in
  recommendation.
\newblock In \emph{Proceedings of the ACM Web Conference 2023}, 2023.
\newblock \doi{10.1145/3543507.3583244}.

\bibitem[Luo et~al.(2023)Luo, Yang, Meng, Li, Zhou, and Zhang]{luo2023}
Yun Luo, Zhicheng Yang, Fandong Meng, Yafu Li, Jie Zhou, and Yue Zhang.
\newblock An empirical study of catastrophic forgetting in large language
  models during continual fine-tuning.
\newblock \emph{arXiv preprint arXiv:2308.08747}, 2023.

\bibitem[Meng et~al.(2022)Meng, Bau, Andonian, and Belinkov]{meng2022rome}
Kevin Meng, David Bau, Alex Andonian, and Yonatan Belinkov.
\newblock Locating and editing factual associations in {GPT}.
\newblock In \emph{Advances in Neural Information Processing Systems}, 2022.

\bibitem[Meng et~al.(2023)Meng, Sharma, Andonian, Belinkov, and
  Bau]{meng2023memit}
Kevin Meng, Arnab~Sen Sharma, Alex Andonian, Yonatan Belinkov, and David Bau.
\newblock Mass-editing memory in a transformer.
\newblock In \emph{International Conference on Learning Representations
  (ICLR)}, 2023.

\bibitem[Min et~al.(2023)Min, Krishna, Lyu, Lewis, Yih, Koh, Iyyer,
  Zettlemoyer, and Hajishirzi]{min2023}
Sewon Min, Kalpesh Krishna, Xinxi Lyu, Mike Lewis, Wen-tau Yih, Pang Koh, Mohit
  Iyyer, Luke Zettlemoyer, and Hannaneh Hajishirzi.
\newblock {FActScore}: Fine-grained atomic evaluation of factual precision in
  long form text generation.
\newblock In \emph{Proceedings of the 2023 Conference on Empirical Methods in
  Natural Language Processing}, pages 12076--12100. Association for
  Computational Linguistics, 2023.
\newblock \doi{10.18653/v1/2023.emnlp-main.741}.
\newblock URL \url{https://aclanthology.org/2023.emnlp-main.741/}.

\bibitem[Ni et~al.(2025{\natexlab{a}})Ni, Bi, Guo, Tang, Wu, Han, and
  Cheng]{elical2025}
Shiyu Ni, Keping Bi, Jiafeng Guo, Minghao Tang, Jingtong Wu, Zengxin Han, and
  Xueqi Cheng.
\newblock Annotation-efficient universal honesty alignment.
\newblock \emph{arXiv preprint arXiv:2510.17509}, 2025{\natexlab{a}}.
\newblock \doi{10.48550/arXiv.2510.17509}.

\bibitem[Ni et~al.(2025{\natexlab{b}})Ni, Bi, Guo, Yu, Bi, and Cheng]{c3_2025}
Shiyu Ni, Keping Bi, Jiafeng Guo, Lulu Yu, Baolong Bi, and Xueqi Cheng.
\newblock Towards fully exploiting {LLM} internal states to enhance knowledge
  boundary perception.
\newblock In \emph{Proceedings of the 63rd Annual Meeting of the Association
  for Computational Linguistics (Volume 1: Long Papers)}, pages 24315--24329.
  Association for Computational Linguistics, 2025{\natexlab{b}}.
\newblock \doi{10.18653/v1/2025.acl-long.1184}.
\newblock URL \url{https://aclanthology.org/2025.acl-long.1184/}.

\bibitem[Nichol et~al.(2018)Nichol, Achiam, and Schulman]{nichol2018reptile}
Alex Nichol, Joshua Achiam, and John Schulman.
\newblock On first-order meta-learning algorithms, 2018.
\newblock URL \url{https://arxiv.org/abs/1803.02999}.

\bibitem[Reder and Ritter(1992)]{reder1992}
Lynne~M. Reder and Frank~E. Ritter.
\newblock What determines initial feeling of knowing? {F}amiliarity with
  question terms, not with the answer.
\newblock \emph{Journal of Experimental Psychology: Learning, Memory, and
  Cognition}, 18\penalty0 (3):\penalty0 435--451, 1992.

\bibitem[Shenfeld et~al.(2026)Shenfeld, Damani, H{\"u}botter, and
  Agrawal]{shenfeld2026sdft}
Idan Shenfeld, Mehul Damani, Jonas H{\"u}botter, and Pulkit Agrawal.
\newblock Self-distillation enables continual learning, 2026.
\newblock URL \url{https://arxiv.org/abs/2601.19897}.

\bibitem[Slobodkin et~al.(2023)Slobodkin, Goldman, Caciularu, Dagan, and
  Ravfogel]{slobodkin2023}
Aviv Slobodkin, Omer Goldman, Avi Caciularu, Ido Dagan, and Shauli Ravfogel.
\newblock The curious case of hallucinatory (un)answerability: Finding truths
  in the hidden states of over-confident large language models.
\newblock In \emph{Proceedings of the 2023 Conference on Empirical Methods in
  Natural Language Processing}, pages 3607--3625. Association for Computational
  Linguistics, 2023.
\newblock \doi{10.18653/v1/2023.emnlp-main.220}.
\newblock URL \url{https://aclanthology.org/2023.emnlp-main.220/}.

\bibitem[Steyvers and Peters(2026)]{steyvers2026metacognition}
Mark Steyvers and Megan A.~K. Peters.
\newblock Metacognition and uncertainty communication in humans and large
  language models.
\newblock \emph{Current Directions in Psychological Science}, 35\penalty0
  (3):\penalty0 131--139, 2026.
\newblock \doi{10.1177/09637214251391158}.

\bibitem[Su et~al.(2024)Su, Wang, Ai, Hu, Wu, Zhou, and Liu]{su2024}
Weihang Su, Changyue Wang, Qingyao Ai, Yiran Hu, Zhijing Wu, Yujia Zhou, and
  Yiqun Liu.
\newblock Unsupervised real-time hallucination detection based on the internal
  states of large language models.
\newblock In \emph{Findings of the Association for Computational Linguistics:
  ACL 2024}, pages 14379--14391. Association for Computational Linguistics,
  2024.
\newblock \doi{10.18653/v1/2024.findings-acl.854}.
\newblock URL \url{https://aclanthology.org/2024.findings-acl.854/}.

\bibitem[Team(2024)]{qwen25}
Qwen Team.
\newblock Qwen2.5 technical report.
\newblock \emph{arXiv preprint arXiv:2412.15115}, 2024.
\newblock \doi{10.48550/arXiv.2412.15115}.

\bibitem[Team(2025)]{qwen3}
Qwen Team.
\newblock Qwen3 technical report.
\newblock \emph{arXiv preprint arXiv:2505.09388}, 2025.
\newblock \doi{10.48550/arXiv.2505.09388}.

\bibitem[Tian et~al.(2023)Tian, Mitchell, Zhou, Sharma, Rafailov, Yao, Finn,
  and Manning]{tian2023}
Katherine Tian, Eric Mitchell, Allan Zhou, Archit Sharma, Rafael Rafailov,
  Huaxiu Yao, Chelsea Finn, and Christopher~D. Manning.
\newblock Just ask for calibration: Strategies for eliciting calibrated
  confidence scores from language models fine-tuned with human feedback.
\newblock In \emph{Proceedings of the 2023 Conference on Empirical Methods in
  Natural Language Processing}, pages 5433--5442. Association for Computational
  Linguistics, 2023.
\newblock \doi{10.18653/v1/2023.emnlp-main.330}.
\newblock URL \url{https://aclanthology.org/2023.emnlp-main.330/}.

\bibitem[Wang et~al.(2023{\natexlab{a}})Wang, Wei, Schuurmans, Le, Chi, Narang,
  Chowdhery, and Zhou]{wang2023}
Xuezhi Wang, Jason Wei, Dale Schuurmans, Quoc~V. Le, Ed~H. Chi, Sharan Narang,
  Aakanksha Chowdhery, and Denny Zhou.
\newblock Self-consistency improves chain of thought reasoning in language
  models.
\newblock In \emph{International Conference on Learning Representations
  (ICLR)}, 2023{\natexlab{a}}.

\bibitem[Wang et~al.(2023{\natexlab{b}})Wang, Li, Sun, and Liu]{ren2023}
Yile Wang, Peng Li, Maosong Sun, and Yang Liu.
\newblock Self-knowledge guided retrieval augmentation for large language
  models.
\newblock In \emph{Findings of the Association for Computational Linguistics:
  EMNLP 2023}, pages 10303--10315. Association for Computational Linguistics,
  2023{\natexlab{b}}.
\newblock \doi{10.18653/v1/2023.findings-emnlp.691}.
\newblock URL \url{https://aclanthology.org/2023.findings-emnlp.691/}.

\bibitem[Wei et~al.(2024)Wei, Nguyen, Chung, Jiao, Papay, Glaese, Schulman, and
  Fedus]{openai2024}
Jason Wei, Karina Nguyen, Hyung~Won Chung, Yunxin~Joy Jiao, Spencer Papay,
  Amelia Glaese, John Schulman, and William Fedus.
\newblock Measuring short-form factuality in large language models
  ({SimpleQA}).
\newblock \emph{arXiv preprint arXiv:2411.04368}, 2024.
\newblock \doi{10.48550/arXiv.2411.04368}.

\bibitem[Xiong et~al.(2024)Xiong, Hu, Lu, Li, Fu, He, and Hooi]{xiong2024}
Miao Xiong, Zhiyuan Hu, Xinyang Lu, Yifei Li, Jie Fu, Junxian He, and Bryan
  Hooi.
\newblock Can {LLM}s express their uncertainty? {A}n empirical evaluation of
  confidence elicitation in {LLM}s.
\newblock In \emph{International Conference on Learning Representations}, 2024.
\newblock URL \url{https://openreview.net/forum?id=gjeQKFxFpZ}.

\bibitem[Xu et~al.(2025)Xu, Li, Zhang, Lin, Zhu, Zheng, Wu, Zhao, Xu, and
  Chen]{xu2025amar}
Derong Xu, Xinhang Li, Ziheng Zhang, Zhenxi Lin, Zhihong Zhu, Zhi Zheng, Xian
  Wu, Xiangyu Zhao, Tong Xu, and Enhong Chen.
\newblock Harnessing large language models for knowledge graph question
  answering via adaptive multi-aspect retrieval-augmentation.
\newblock In \emph{Proceedings of the AAAI Conference on Artificial
  Intelligence}, volume~39, pages 25570--25578, 2025.
\newblock \doi{10.1609/aaai.v39i24.34747}.

\bibitem[Yin et~al.(2023)Yin, Sun, Guo, Wu, Qiu, and Huang]{yin2023selfaware}
Zhangyue Yin, Qiushi Sun, Qipeng Guo, Jiawen Wu, Xipeng Qiu, and Xuanjing
  Huang.
\newblock Do large language models know what they don't know?
\newblock In \emph{Findings of the Association for Computational Linguistics:
  ACL 2023}, pages 8653--8665, 2023.
\newblock \doi{10.18653/v1/2023.findings-acl.551}.
\newblock URL \url{https://aclanthology.org/2023.findings-acl.551/}.

\bibitem[Zhang et~al.(2024)Zhang, Diao, Lin, Fung, Lian, Wang, Chen, Ji, and
  Zhang]{zhang2023rtuning}
Hanning Zhang, Shizhe Diao, Yong Lin, Yi~R. Fung, Qing Lian, Xingyao Wang,
  Yangyi Chen, Heng Ji, and Tong Zhang.
\newblock {R-Tuning}: Instructing large language models to say `{I} don't
  know'.
\newblock In \emph{Proceedings of the 2024 Conference of the North American
  Chapter of the Association for Computational Linguistics: Human Language
  Technologies (Volume 1: Long Papers)}, pages 7113--7139. Association for
  Computational Linguistics, 2024.
\newblock \doi{10.18653/v1/2024.naacl-long.394}.
\newblock URL \url{https://aclanthology.org/2024.naacl-long.394/}.

\bibitem[Zhang et~al.(2025)Zhang, Li, Zhang, Jia, Wang, Guo, Liu, and
  Zhao]{zhang2025deepresearch}
Wenlin Zhang, Xiaopeng Li, Yingyi Zhang, Pengyue Jia, Yichao Wang, Huifeng Guo,
  Yong Liu, and Xiangyu Zhao.
\newblock Deep research: A survey of autonomous research agents.
\newblock \emph{arXiv preprint arXiv:2508.12752}, 2025.
\newblock \doi{10.48550/arXiv.2508.12752}.

\bibitem[Zhang et~al.(2023)Zhang, Zhao, Liu, Zhang, Ma, Wang, Zhao, Wang, and
  Liu]{zhang2023promptst}
Zijian Zhang, Xiangyu Zhao, Qidong Liu, Chunxu Zhang, Qian Ma, Wanyu Wang,
  Hongwei Zhao, Yiqi Wang, and Zitao Liu.
\newblock {PromptST}: Prompt-enhanced spatio-temporal multi-attribute
  prediction.
\newblock In \emph{Proceedings of the 32nd ACM International Conference on
  Information and Knowledge Management}, pages 3195--3205, 2023.

\bibitem[Zhao et~al.(2018{\natexlab{a}})Zhao, Xia, Zhang, Ding, Yin, and
  Tang]{zhao2018pagewise}
Xiangyu Zhao, Long Xia, Liang Zhang, Zhuoye Ding, Dawei Yin, and Jiliang Tang.
\newblock Deep reinforcement learning for page-wise recommendations.
\newblock In \emph{Proceedings of the 12th ACM Conference on Recommender
  Systems}, pages 95--103, 2018{\natexlab{a}}.
\newblock \doi{10.1145/3240323.3240374}.

\bibitem[Zhao et~al.(2018{\natexlab{b}})Zhao, Zhang, Ding, Xia, Tang, and
  Yin]{zhao2018deers}
Xiangyu Zhao, Liang Zhang, Zhuoye Ding, Long Xia, Jiliang Tang, and Dawei Yin.
\newblock Recommendations with negative feedback via pairwise deep
  reinforcement learning.
\newblock In \emph{Proceedings of the 24th ACM SIGKDD International Conference
  on Knowledge Discovery \& Data Mining}, 2018{\natexlab{b}}.

\bibitem[Zhao et~al.(2021)Zhao, Xia, Zou, Liu, Yin, and Tang]{zhao2021usersim}
Xiangyu Zhao, Long Xia, Lixin Zou, Hui Liu, Dawei Yin, and Jiliang Tang.
\newblock {UserSim}: User simulation via supervised generative adversarial
  network.
\newblock In \emph{Proceedings of the Web Conference 2021}, pages 3582--3589,
  2021.
\newblock \doi{10.1145/3442381.3450125}.

\bibitem[Zheng et~al.(2023)Zheng, Chiang, Sheng, Zhuang, Wu, Zhuang, Lin, Li,
  Li, Xing, Zhang, Gonzalez, and Stoica]{zheng2023judging}
Lianmin Zheng, Wei-Lin Chiang, Ying Sheng, Siyuan Zhuang, Zhanghao Wu, Yonghao
  Zhuang, Zi~Lin, Zhuohan Li, Dacheng Li, Eric~P. Xing, Hao Zhang, Joseph~E.
  Gonzalez, and Ion Stoica.
\newblock Judging {LLM}-as-a-judge with {MT-Bench} and chatbot arena.
\newblock In \emph{Advances in Neural Information Processing Systems}, 2023.

\bibitem[Zhou et~al.(2025)Zhou, Xie, Gong, Wen, Zhao, Xuan, and
  Yang]{zhou2025gdaug}
Jiajun Zhou, Chenxuan Xie, Shengbo Gong, Zhenyu Wen, Xiangyu Zhao, Qi~Xuan, and
  Xiaoniu Yang.
\newblock Data augmentation on graphs: A technical survey.
\newblock \emph{ACM Computing Surveys}, 57\penalty0 (11):\penalty0 274, 2025.
\newblock \doi{10.1145/3732282}.

\bibitem[Zhu et~al.(2025{\natexlab{a}})Zhu, Jiang, Wu, Ma, Song, Bai, Lin, Wu,
  and He]{chen2025grait}
Runchuan Zhu, Zinco Jiang, Jiang Wu, Zhipeng Ma, Jiahe Song, Fengshuo Bai,
  Dahua Lin, Lijun Wu, and Conghui He.
\newblock {GRAIT}: Gradient-driven refusal-aware instruction tuning for
  effective hallucination mitigation.
\newblock In \emph{Findings of the Association for Computational Linguistics:
  NAACL 2025}, pages 4006--4021. Association for Computational Linguistics,
  2025{\natexlab{a}}.
\newblock \doi{10.18653/v1/2025.findings-naacl.223}.
\newblock URL \url{https://aclanthology.org/2025.findings-naacl.223/}.

\bibitem[Zhu et~al.(2025{\natexlab{b}})Zhu, Ma, Wu, Gao, Wang, Lin, and
  He]{wang2024craft}
Runchuan Zhu, Zhipeng Ma, Jiang Wu, Junyuan Gao, Jiaqi Wang, Dahua Lin, and
  Conghui He.
\newblock Utilize the flow before stepping into the same river twice: Certainty
  represented knowledge flow for refusal-aware instruction tuning.
\newblock In \emph{Proceedings of the AAAI Conference on Artificial
  Intelligence}, volume~39, pages 26157--26165, 2025{\natexlab{b}}.
\newblock \doi{10.1609/aaai.v39i24.34812}.

\end{thebibliography}

\appendix
\raggedbottom
\section{Limitations and future work}
\label{app:limitations}

We evaluate the framework on short-answer factual QA with
gradient-based updates, including three sequential update rounds.
The MEMIT experiment in Appendix~\ref{app:additional-validation} tests
desynchronization under one non-gradient editor, not whether our full
framework repairs that editor. Longer update sequences, procedural
tasks such as code generation and mathematical proof, and backbones
whose baseline FoK signal is near chance remain outside the tested
scope. Strong unadapted estimators remain competitive without an update.

The cue analyses do not include a matched
pretrained-model control, so they cannot establish whether training
induces cue sensitivity or preserves sensitivity already in the
backbone. Causal recovery locations depend on the prompt, token
positions, and intervention; they do not establish a uniquely localized
or human-equivalent metacognitive mechanism. More broadly, we use FoK in a functional sense, as answerability judgments modeled after human FoK, without implying subjective experience. Seed variation covers
Stages 1--2 and FGPO on the main TriviaQA setting, not every preliminary
or cross-dataset experiment.

\section{Broader impact}
\label{app:broader-impact}

This work targets reliability after LLM knowledge updates by making
answerability judgments track the updated model state. The intended
positive impact is fewer unsupported answers after fine-tuning or
continual updates. A possible negative impact is that better
answerability judgments could increase trust in systems whose
deployment still depends on data quality, monitoring, and
domain-specific safety constraints. The method is not a content-safety
system on its own and should be combined with deployment-specific safeguards before release.

\section{Additional experimental settings}
\label{app:hyper}

\paragraph{Data construction.}
The main experiments use TriviaQA as the knowledge-update corpus and the
in-domain QA benchmark. The update set contains 5,000 questions, and the
held-out TriviaQA test split contains 1,000 questions. For OP-SFT, we sample
multiple responses from the current answer model for each update question,
repair the answer-relevant span when the sampled response is wrong, and train
only the corrected answer span. The reported main setting uses 32 sampled
responses per question. General capability is evaluated separately on MMLU and
ARC-Easy, using the same generator whose TriviaQA result is reported in the
main table.

\paragraph{Evaluation protocol.}
Answer accuracy on TriviaQA is the fraction of generated responses that match the gold answer, with surface-form equivalence checked by Qwen2.5-32B-Instruct~\citep{qwen25} used as an LLM judge~\citep{zheng2023judging}. MMLU and ARC-Easy use their standard multiple-choice accuracy. Generation and judging are run with vLLM in bfloat16 precision; the 32B judge uses tensor parallelism across the available GPUs. FoK metrics use the model's pre-answer Yes/No judgment score: AUROC is computed from the normalized probability of the Yes token, and illusion, omission, and Bias are computed from the thresholded Yes/No prediction. Since illusion and omission are conditional error rates in $[0,1]$, the signed Bias score lies in $[-1,1]$. Positive values mean illusion errors dominate, and negative values mean omission errors dominate.

\paragraph{Baseline protocol.}
All FoK estimators are evaluated under the answer model used by that row. Base
rows use the unadapted model; Vanilla K-SFT rows use the knowledge-updated
model; Ours uses the combined knowledge and FoK adapters. Verbal-0 and Verbal-10 ask the
model for a direct confidence judgment, Prob and N-Prob use sequence
likelihood, Consis-Lex and Consis-Sem use multiple sampled responses, semantic
entropy clusters sampled responses by meaning, and EliCal uses the released
EliCal adapter whenever its public checkpoint is available for the backbone.
Unlike estimators that only score fixed answers, loading the EliCal adapter
can change answering behavior and answer accuracy. The fixed-answer EliCal
transfer protocol in Appendix~\ref{app:additional-validation} is reported
separately.

\paragraph{Training protocol.}
All trainable components are implemented as LoRA adapters on top of
Qwen2.5-7B-Instruct. Vanilla K-SFT is trained with standard supervised
fine-tuning on the same update questions. OP-SFT trains the knowledge adapter
from repaired sampled responses. K$\Delta$-FT freezes the OP-SFT knowledge
checkpoints and trains a FoK adapter against labels collected from those
knowledge states. FGPO then uses the knowledge and FoK adapters jointly for
response optimization. Table~\ref{tab:hyper} lists the full training
hyperparameters.

\paragraph{Compute resources.}
The main experiments were run on a single multi-GPU server. Training used
multi-GPU DDP with DeepSpeed ZeRO-2. Generation and judging used vLLM in
bfloat16 precision, and the Qwen2.5-32B judge used tensor parallelism across
the available GPUs. We report the parallelism and training configuration
needed for reproduction, but did not log exact wall-clock time for every
preliminary and failed run.

\begin{table}[htbp]
  \centering
  \caption{Training hyperparameters for the main experiments.}
  \label{tab:hyper}
  
  \begin{tabular}{@{}ll@{}}
    \toprule
    Parameter & Value \\
    \midrule
    Backbone & Qwen2.5-7B-Instruct \\
    Training framework & LLaMA Factory \\
    Adapter type & LoRA, rank 16, alpha 32, dropout 0 \\
    Parallelism & Multi-GPU DDP with DeepSpeed ZeRO-2 \\
    Precision & bfloat16 \\
    Vanilla K-SFT LR / epochs & $10^{-4}$ / 5 \\
    OP-SFT sampled responses / epochs & 32 / 1 \\
    OP-SFT LR & $2\times 10^{-5}$ \\
    K$\Delta$-FT LR / epochs & $10^{-4}$ / 3 \\
    FGPO LR / epochs & $10^{-6}$ / 1 \\
    Per-device batch size & 1 \\
    Gradient accumulation & 1 \\
    Schedule         & Cosine with 0.03 warmup ratio \\
    FoK score    & normalized \texttt{Yes}/\texttt{No} token probability \\
    \bottomrule
  \end{tabular}
\end{table}

\paragraph{FGPO reward values.}
Table~\ref{tab:fgpo-reward} lists the implemented FGPO rewards. The
base reward gives $R_{\mathrm{FoK}}+R_{\mathrm{ans}}$ for each
combination of answerability target, judgment, and response outcome.
An output from which no yes/no judgment can be parsed receives $-1.0$
in place of the base reward. A further $+0.1$ is added when the output
follows the full \texttt{<think>}/\texttt{<answer>}/\texttt{<response>}
format, $+0.1$ when generation ends with the EOS token and $-0.2$
otherwise, and the length term uses $\lambda=0.001$ with $|h|$ measured
in characters of the reasoning trace.

\begin{table}[htbp]
  \centering
  \caption{Base FGPO reward $R_{\mathrm{FoK}}+R_{\mathrm{ans}}$ by
  target $Y$, judgment $j$, and response outcome.}
  \label{tab:fgpo-reward}
  \begin{tabular}{@{}cclc@{}}
    \toprule
    $Y$ & $j$ & Response outcome & Base reward \\
    \midrule
    1 & yes & correct answer & $+2.0$ \\
    1 & yes & wrong answer & $-1.5$ \\
    1 & no & --- & $-1.0$ \\
    0 & no & valid explanation & $+0.5$ \\
    0 & no & invalid explanation & $-0.5$ \\
    0 & yes & correct answer & $-0.5$ \\
    0 & yes & wrong answer & $-2.0$ \\
    \bottomrule
  \end{tabular}
\end{table}

\paragraph{Existing assets and terms of use.}
Table~\ref{tab:asset-licenses} lists the external datasets, models, and code
packages used in the experiments. We use each asset only for research
evaluation or fine-tuning under the terms specified by the original publisher.

\begin{table}[htbp]
  \centering
  \caption{External assets used in the experiments and their publisher-specified
  licenses or terms of use.}
  \label{tab:asset-licenses}
  \resizebox{\textwidth}{!}{%
\begin{tabular}{@{}lll@{}}
    \toprule
    Asset & Use in this paper & License or terms \\
    \midrule
    TriviaQA & \begin{tabular}[c]{@{}l@{}}Knowledge-update corpus\\and in-domain QA benchmark\end{tabular} & Apache-2.0 (\href{https://github.com/mandarjoshi90/triviaqa/blob/master/LICENSE}{source}) \\
    MMLU & General capability benchmark & MIT License (\href{https://github.com/hendrycks/test/blob/master/LICENSE}{source}) \\
    ARC-Easy & General capability benchmark & CC BY-SA 4.0 (\href{https://huggingface.co/datasets/allenai/ai2_arc}{source}) \\
\begin{tabular}[c]{@{}l@{}}Qwen2.5-7B-Instruct and\\Qwen2.5-32B-Instruct\end{tabular} & \begin{tabular}[c]{@{}l@{}}Main answer model\\and LLM judge\end{tabular} & Apache-2.0 (Qwen \href{https://huggingface.co/Qwen/Qwen2.5-7B-Instruct}{7B} and \href{https://huggingface.co/Qwen/Qwen2.5-32B-Instruct}{32B} model cards) \\
    Qwen3-8B & \begin{tabular}[c]{@{}l@{}}Thinking-mode backbone for\\the selective-prediction evaluation\end{tabular} & Apache-2.0 (Qwen \href{https://huggingface.co/Qwen/Qwen3-8B}{model card}) \\
\begin{tabular}[c]{@{}l@{}}LLaMA-3.1 and\\LLaMA-2 checkpoints\end{tabular} & \begin{tabular}[c]{@{}l@{}}Supplementary FoK synchronization\\results in Appendix~\ref{app:full_table}\end{tabular} & \begin{tabular}[c]{@{}l@{}}Meta Llama 3.1 Community License\\and Llama 2 Community License (\href{https://huggingface.co/meta-llama/Llama-3.1-8B/blob/main/LICENSE}{3.1}, \href{https://huggingface.co/meta-llama/Llama-2-7b/blob/main/LICENSE.txt}{2})\end{tabular} \\
    LLaMA Factory & LoRA fine-tuning framework & Apache-2.0 (\href{https://github.com/hiyouga/LLaMA-Factory/blob/main/LICENSE}{source}) \\
    vLLM & \begin{tabular}[c]{@{}l@{}}Batched generation\\and LLM judging\end{tabular} & Apache-2.0 (\href{https://github.com/vllm-project/vllm/blob/main/LICENSE}{source}) \\
\begin{tabular}[c]{@{}l@{}}EleutherAI\\lm-evaluation-harness\end{tabular} & MMLU and ARC-Easy evaluation & MIT License (\href{https://github.com/EleutherAI/lm-evaluation-harness/blob/main/LICENSE.md}{source}) \\
    EliCal official adapter & FoK baseline & Publisher model-card terms (\href{https://huggingface.co/Shiyunee/Honest-Qwen2.5-7B-Instruct}{source}) \\
    \bottomrule
  \end{tabular}
  }
\end{table}

\FloatBarrier
\section{Causal tracing and weight-delta supplementary figures}
\label{app:f1-causal-tracing}

\begin{figure}[H]
  \centering
  \includegraphics[width=0.9\linewidth]{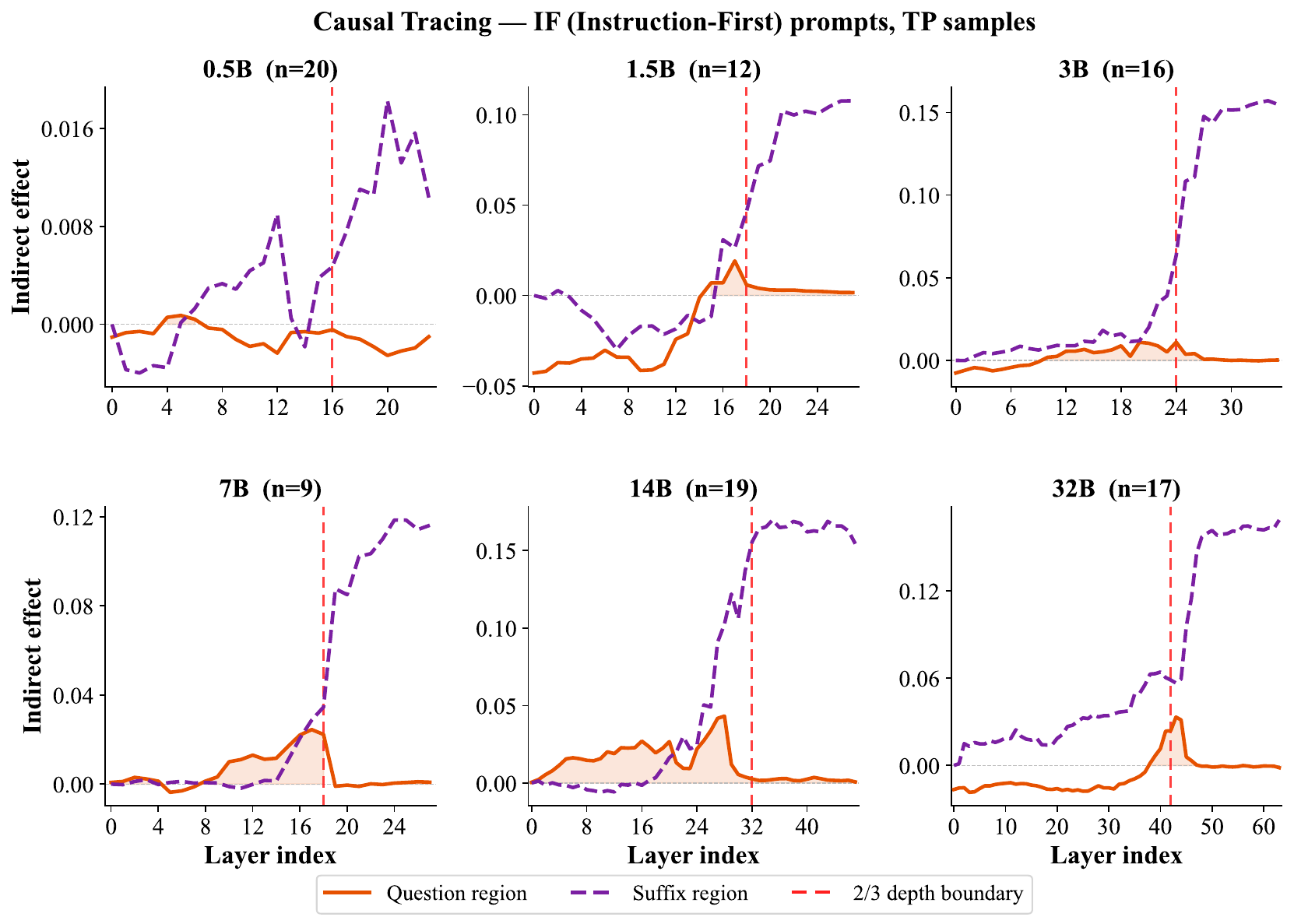}
  \caption{Per-layer indirect effect on true-positive (TP) samples for the
  instruction-first (IF) prompt across six Qwen2.5 sizes. The solid orange
  curve shows the question-region effect, the dashed purple curve shows the
  suffix-region effect, and the red dashed line marks two-thirds of model
  depth. The question-region effect concentrates in early and middle
  layers, whereas the suffix-region effect rises mainly after this boundary
  at every tested scale.}
  \label{fig:f1-if-curves}
\end{figure}

\begin{figure}[htbp]
  \centering
  \includegraphics[width=\linewidth]{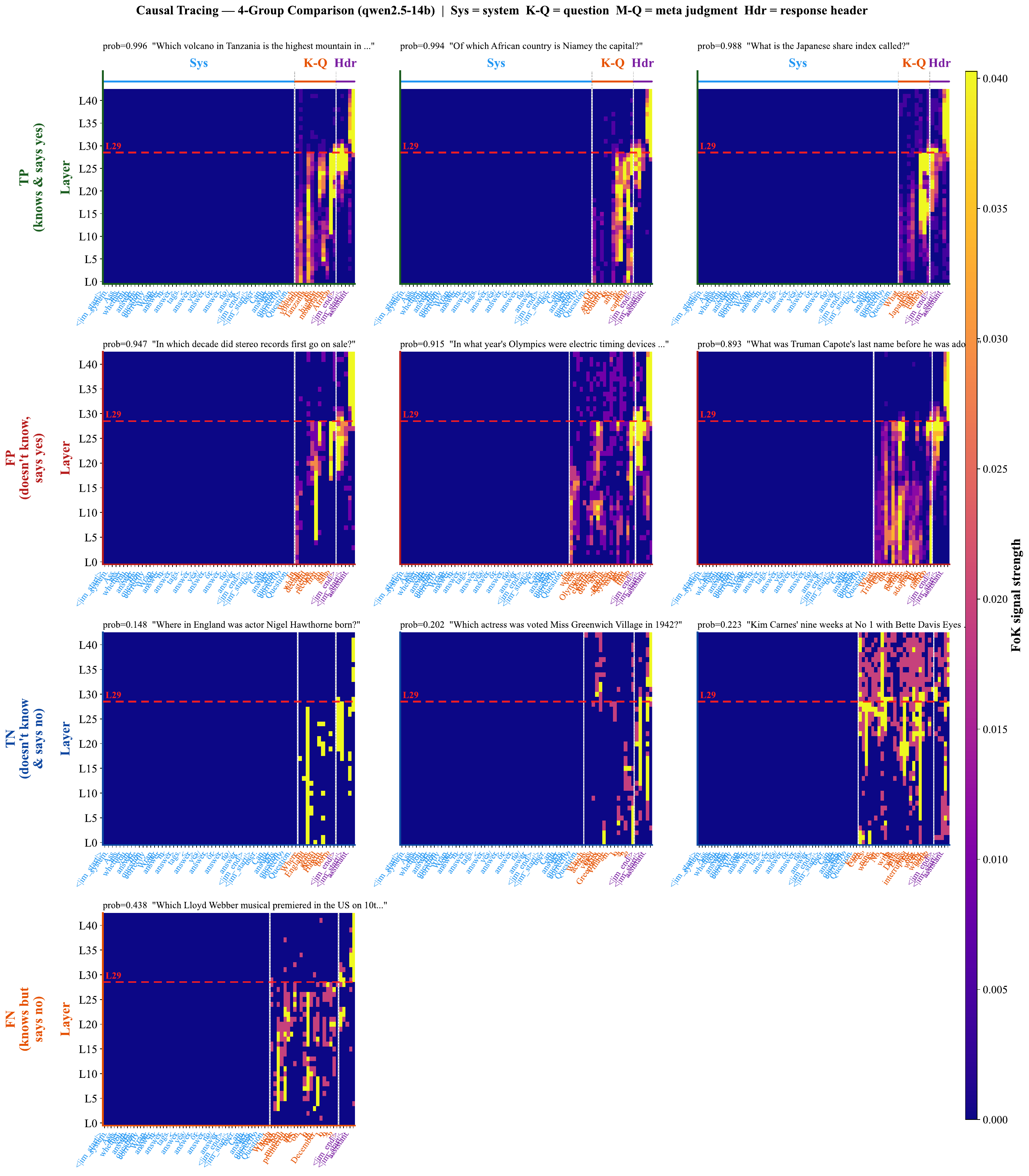}
  \caption{ROME-style causal tracing on Qwen2.5-14B, broken down by
  outcome group. Rows: TP (answers correctly and judges yes), FP (answers
  incorrectly but judges yes), TN (answers incorrectly and judges no),
  and FN (answers correctly but judges no). Each row shows three samples,
  except FN, which has only one sample with $p>0.4$. Each cell
  is the layer\,$\times$\,token restoration heatmap; tokens are grouped
  into Sys (system), K-Q (knowledge question), and Hdr (response
  header). The red dashed line at layer 29 of 48 (about 60\% of depth) marks the depth
  boundary used in our analysis. The suffix-region (Hdr) signal appears
  mainly in deep layers across all four outcome groups, suggesting that
  the late FoK signal does not simply track answer correctness.}
  \label{fig:f1-4group}
\end{figure}

\begin{figure}[htbp]
  \centering
  \includegraphics[width=0.7\linewidth]{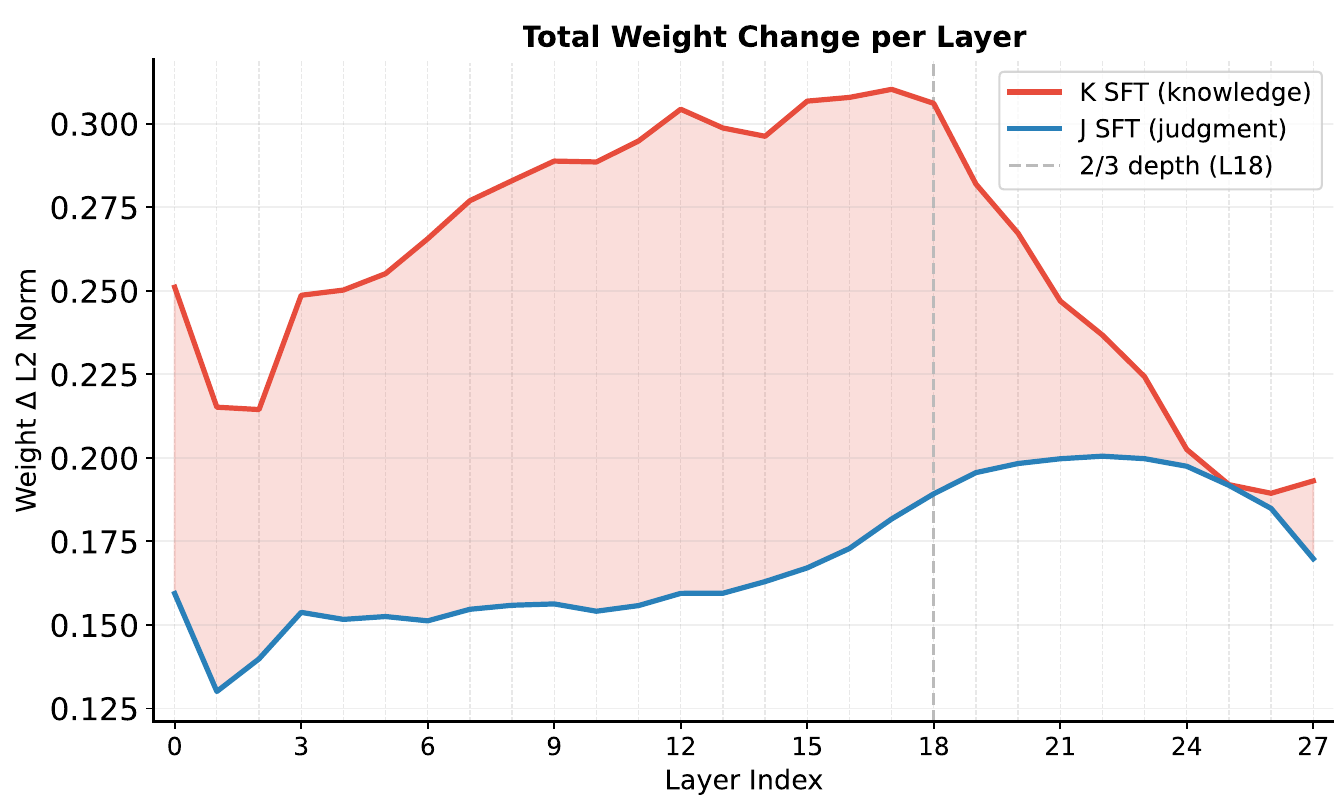}
  \caption{Layer-wise weight delta $\lVert\Delta W_\ell\rVert_F$ after
  one epoch of K-SFT (red) vs.\ J-SFT (blue) on Qwen2.5-7B (full SFT
  from base, same data budget). K-SFT changes concentrate in
  shallow/mid layers, while J-SFT changes remain small until the deep
  layers. This weight-side pattern is consistent with the depth separation
  observed in causal tracing.}
  \label{fig:f1-weight-delta}
\end{figure}

\FloatBarrier

\section{FoK dependency on knowledge}
\label{app:f2-supplementary}

\begin{figure}[H]
  \centering
  \begin{minipage}[t]{0.48\linewidth}
    \centering
    \vspace{0pt}
    \includegraphics[width=\linewidth]{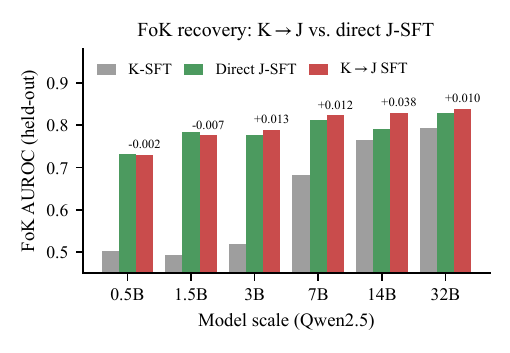}
    \subcaption{FoK recovery. Numbers above the K$\to$J bars give the
    gap to direct J-SFT. K$\to$J recovers FoK at every scale, stays
    within 0.007 of direct J-SFT at 0.5B and 1.5B, and exceeds it from
    3B upward.}
    \label{fig:f2-recovery}
  \end{minipage}\hfill
  \begin{minipage}[t]{0.48\linewidth}
    \centering
    \vspace{0pt}
    \includegraphics[width=\linewidth]{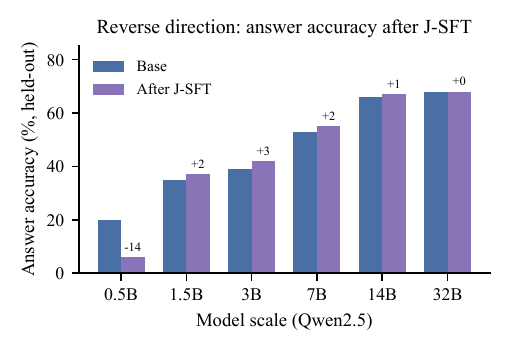}
    \subcaption{Reverse direction. Answer accuracy before and after
    J-SFT. From 1.5B upward, J-SFT changes accuracy by at most 3
    points; at 0.5B it lowers accuracy by 14 points.}
    \label{fig:f2-reverse}
  \end{minipage}
  \caption{FoK dependence on updated knowledge, from the same held-out
  TriviaQA runs as Table~\ref{tab:f2-knowledge-judgment-coupling}. Left: judgment
  training after a knowledge update restores FoK at every scale. Right:
  judgment training alone leaves answer accuracy largely unchanged
  except at 0.5B.}
  \label{fig:f2-supplementary}
\end{figure}

\section{FoK judgment along OP-SFT training}
\label{app:trajectory}

\begin{figure}[H]
  \centering
  \includegraphics[width=\linewidth]{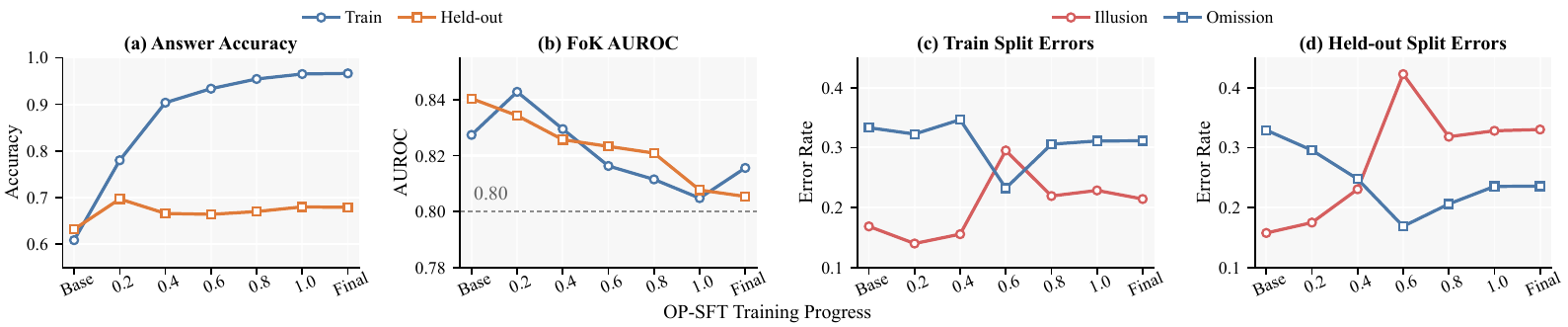}
  \caption{In a separate small-scale study, we evaluate a fixed, trained
  K$\Delta$-FT FoK channel across successive OP-SFT knowledge checkpoints.
  We report answer accuracy, FoK AUROC, and thresholded illusion and omission errors.}
  \label{fig:stage2-kplusfok-trajectory}
\end{figure}

Figure~\ref{fig:stage2-kplusfok-trajectory} tests whether the trained FoK
channel only works at the final OP-SFT checkpoint. Answer accuracy changes
substantially along the OP-SFT trajectory, while FoK AUROC remains above 0.80 on
both splits for every checkpoint. The thresholded errors show a calibration
shift: omission is larger near the base model, whereas illusion grows on later
held-out checkpoints. Thus K$\Delta$-FT provides a stable ranking signal across
the knowledge trajectory, but threshold calibration still depends on the underlying knowledge state of the model.

\section{FoK synchronization results across model scale and family}
\label{app:full_table}

\begin{table}[H]
  \centering
  \caption{Qwen2.5 size scaling in a dedicated setup
  (5,000 update and 1,000 held-out questions).}
  \label{tab:full}
  \resizebox{\textwidth}{!}{%
\begin{tabular}{@{}lcccccccccc@{}}
    \toprule
    & \multicolumn{2}{c}{Base} & \multicolumn{4}{c}{K-SFT} & J-SFT & \multicolumn{3}{c}{K$\to$J SFT} \\
    \cmidrule(lr){2-3}\cmidrule(lr){4-7}\cmidrule(lr){8-8}\cmidrule(lr){9-11}
    Model & EM & AUROC & EM & AUROC & Illu. & Om. & AUROC & EM & AUROC & Illu. \\
    \midrule
    0.5B & 4.2\%  & 0.523 & 16.9\% & 0.502 & 0\%  & 99\% & 0.801 & 13.1\% & 0.763 & 3\%  \\
    1.5B & 30.9\% & 0.527 & 35.5\% & 0.500 & 66\% & 48\% & 0.824 & 34.8\% & 0.815 & 28\% \\
    3B   & 35.4\% & 0.724 & 45.7\% & 0.519 & 24\% & 80\% & 0.830 & 45.4\% & 0.822 & 40\% \\
    7B   & 47.2\% & 0.735 & 55.8\% & 0.592 & 28\% & 51\% & 0.855 & 55.1\% & 0.833 & 82\% \\
    14B  & 59.8\% & 0.756 & 65.2\% & 0.767 & 73\% & 9\%  & 0.850 & 66.1\% & 0.814 & 85\% \\
    32B  & 61.2\% & 0.812 & 66.8\% & 0.792 & 80\% & 5\%  & 0.844 & 66.8\% & 0.831 & 91\% \\
    \bottomrule
  \end{tabular}
  }
\end{table}

\begin{table}[H]
  \centering
  \caption{Cross-family FoK synchronization results with
  5,000 update and 1,000 held-out questions.}
  \label{tab:cross_family}
  \resizebox{\textwidth}{!}{%
\begin{tabular}{@{}lcccccc@{}}
    \toprule
    Model & \begin{tabular}[c]{@{}c@{}}Base\\AUROC\end{tabular} & \begin{tabular}[c]{@{}c@{}}K-SFT\\AUROC\end{tabular} & \begin{tabular}[c]{@{}c@{}}J-SFT\\AUROC\end{tabular} & \begin{tabular}[c]{@{}c@{}}K$\to$J SFT\\AUROC\end{tabular} & \begin{tabular}[c]{@{}c@{}}K$\to$J SFT\\EM\end{tabular} & Note \\
    \midrule
    Qwen2.5-7B & 0.735 & 0.592 & 0.855 & 0.833 & 55.1\% & Reference \\
    \begin{tabular}[c]{@{}l@{}}LLaMA-3.1-8B\\(LR $=2\times10^{-6}$)\end{tabular} & 0.646 & 0.503 & 0.786 & 0.836 & 8.6\%  & K$\to$J forgetting \\
    \begin{tabular}[c]{@{}l@{}}LLaMA-3.1-8B\\(LR $=10^{-6}$)\end{tabular} & 0.646 & 0.526 & 0.782 & 0.749 & 73.5\% & LR corrected \\
    LLaMA-2-7B             & 0.520 & 0.500 & 0.500 & 0.500 & 63.5\% & No FoK improvement \\
    LLaMA-2-13B            & 0.525 & 0.500 & 0.500 & 0.500 & 70.1\% & No FoK improvement \\
    \bottomrule
  \end{tabular}
  }
\end{table}

\paragraph{Knowledge training and FoK degradation.}
Consistent with the main-text findings, the K-SFT results in
Table~\ref{tab:full} show that knowledge training can improve answer
accuracy while impairing FoK. An update optimized
for answering alone need not preserve the model's ability to distinguish
its correct answers from errors, motivating explicit FoK synchronization
rather than assuming that knowledge gains automatically improve self-judgment.

\paragraph{LLaMA-2 negative result.}
The LLaMA-2 results reveal a distinction between acquiring knowledge and
judging answerability. Although answer accuracy improves, FoK remains at
chance after both knowledge and judgment training
(Table~\ref{tab:cross_family}). This suggests that correctness supervision
alone is insufficient to elicit a discriminative FoK signal in this setting:
the model can learn answers without learning to distinguish when its own
answers are correct.

\section{Additional validation}
\label{app:additional-validation}

We conduct additional small-scale experiments, each with its own training
and evaluation setup, to examine different aspects of FoK behavior.

\subsection{Accuracy-preserving updates and sequential knowledge changes}

\paragraph{OP-SFT comparison.}
Table~\ref{tab:opsft-full} compares FoK estimators starting from the
same OP-SFT knowledge checkpoint. Adding a FoK adapter changes the
combined model state, and the K$\Delta$-FT row therefore has a different
answer accuracy.
For Base, EliCal uses its released adapter, which is trained extensively
and therefore changes answering behavior. When combined with OP-SFT,
EliCal uses an adapter with only a small amount of training, so answer
accuracy is unchanged.

\begin{table}[H]
\centering
\caption{FoK evaluation before and after OP-SFT on the 1,000-question held-out TriviaQA split. All OP-SFT rows start from the same knowledge checkpoint.}
\label{tab:opsft-full}
\begin{tabular}{@{}llcc@{}}
\toprule
Answer model & FoK estimator & Test Acc. & Test AUROC\\
\midrule
Base & Direct judgment & 0.672 & 0.688\\
Base & Prob & 0.672 & 0.730\\
Base & EliCal & 0.622 & 0.810\\
\midrule
OP-SFT & Direct judgment & 0.679 & 0.603\\
OP-SFT & Prob & 0.679 & 0.571\\
OP-SFT & EliCal & 0.679 & 0.618\\
OP-SFT + FoK adapter & K$\Delta$-FT & 0.699 & 0.770\\
\bottomrule
\end{tabular}
\end{table}

\paragraph{Three sequential updates.}
We divide 5,000 TriviaQA update questions into three disjoint shards.
Round $i$ applies OP-SFT with on-policy rollouts to the previous
round's model using shard $i$. The frozen FoK channel $\varphi_1$ is
trained after round 1 and kept fixed thereafter. The adaptive channel
$\varphi_i$ is trained with K$\Delta$-FT on each round's knowledge
trajectory. All held-out columns use the same 1,000 questions.
Table~\ref{tab:sequential-full} reports before/after answer accuracy
on the \emph{new shard for that round}.
The adaptive channel maintains discrimination and reduces illusion in
later rounds, but held-out answer accuracy remains below Base.

\begin{table}[H]
\centering
\caption{Three sequential TriviaQA updates. Illusion is the conditional rate of predicting Yes on questions the current answering model gets wrong.}
\label{tab:sequential-full}
\begin{tabular}{@{}lcccccc@{}}
\toprule
 & New-shard & Test & \multicolumn{2}{c}{$\varphi_1$ (frozen)} & \multicolumn{2}{c}{$\varphi_i$ (adaptive)}\\
\cmidrule(lr){4-5}\cmidrule(lr){6-7}
State & Acc. before $\to$ after & Acc. & AUROC & Illusion & AUROC & Illusion\\
\midrule
Base & --- & 0.672 & --- & --- & --- & ---\\
Round 1 & $0.575\to0.972$ & 0.659 & 0.798 & 0.455 & 0.798 & 0.455\\
Round 2 & $0.676\to0.938$ & 0.644 & 0.776 & 0.688 & 0.800 & 0.202\\
Round 3 & $0.631\to0.938$ & 0.661 & 0.754 & 0.752 & 0.797 & 0.195\\
\bottomrule
\end{tabular}
\end{table}

\subsection{Cross-dataset evaluation and a non-gradient update}

\paragraph{Zero-shot transfer.}
We evaluate the TriviaQA-trained systems on SelfAware (3,369 questions),
Natural Questions (3,610), and WebQuestions (2,032), without additional
training (Table~\ref{tab:cross-dataset}). FoK targets each evaluated
model's own answer correctness. In SelfAware, unanswerable questions
count as incorrect answers; this adapts the original testbed to
model-specific answerability rather than using its answerability
annotation as a fixed target. In this transfer experiment, EliCal
scores the fixed answers of the corresponding Base or OP-SFT generator,
so each paired row has the same answer accuracy. This differs from the
adapter-based answering evaluation of Table~\ref{tab:opsft-full}.

\begin{table}[H]
\centering
\caption{Zero-shot transfer from TriviaQA. SA: SelfAware; NQ: Natural Questions; WebQ: WebQuestions. Acc. measures answer correctness; AUROC measures FoK discrimination.}
\label{tab:cross-dataset}
\begin{tabular}{@{}lcccccc@{}}
\toprule
 & \multicolumn{2}{c}{SA} & \multicolumn{2}{c}{NQ} & \multicolumn{2}{c}{WebQ}\\
\cmidrule(lr){2-3}\cmidrule(lr){4-5}\cmidrule(lr){6-7}
System & Acc. & AUROC & Acc. & AUROC & Acc. & AUROC\\
\midrule
Base & 0.282 & 0.768 & 0.376 & 0.665 & 0.527 & 0.597\\
Base + EliCal & 0.282 & 0.702 & 0.376 & 0.735 & 0.527 & 0.636\\
Vanilla K-SFT & 0.187 & 0.569 & 0.191 & 0.571 & 0.224 & 0.519\\
OP-SFT + EliCal & 0.305 & 0.590 & 0.381 & 0.579 & 0.539 & 0.515\\
Ours & 0.305 & 0.810 & 0.381 & 0.733 & 0.539 & 0.644\\
\bottomrule
\end{tabular}
\end{table}

\begin{table}[H]
\centering
\caption{SelfAware detail for the same zero-shot systems. TPR is the rate of predicting Yes on correctly answered questions; TNR is the rate of predicting No on incorrectly answered questions.}
\label{tab:selfaware-detail}
\begin{tabular}{@{}lcccc@{}}
\toprule
System & AUROC & TPR & TNR & Answer Acc.\\
\midrule
Base & 0.768 & 0.388 & 0.898 & 0.282\\
Vanilla K-SFT & 0.569 & 0.302 & 0.760 & 0.187\\
Ours & 0.810 & 0.765 & 0.701 & 0.305\\
\bottomrule
\end{tabular}
\end{table}

\paragraph{Natural Questions replication.}
Separately from zero-shot transfer, we run the full pipeline on a
5,000-update / 1,000-held-out Natural Questions split
(Table~\ref{tab:nq-replication}). The resulting update-shard accuracy
is 0.925. This replicates the knowledge-update comparison
on a second factual-QA dataset.

\begin{table}[H]
\centering
\caption{Full-pipeline replication on Natural Questions. This held-out split differs from the zero-shot population in Table~\ref{tab:cross-dataset}.}
\label{tab:nq-replication}
\begin{tabular}{@{}lcc@{}}
\toprule
System & Held-out Acc. & FoK AUROC\\
\midrule
Base & 0.319 & 0.629\\
Vanilla K-SFT & 0.199 & 0.585\\
Ours & 0.366 & 0.647\\
\bottomrule
\end{tabular}
\end{table}

\paragraph{MEMIT.}
We apply one MEMIT batch edit~\citep{meng2023memit} to 200 facts in
the zsRE setting~\citep{levy2017zsre}. Table~\ref{tab:memit-fok} measures the edited facts
and the same 1,000 held-out TriviaQA questions before and after editing.
Edited-fact accuracy rises while FoK AUROC decreases. Knowledge and FoK thus also
desynchronize under a non-gradient operator.

\begin{table}[H]
\centering
\caption{Knowledge and functional FoK before and after a MEMIT edit. Train denotes 200 edited zsRE facts; Test denotes 1,000 held-out TriviaQA questions.}
\label{tab:memit-fok}
\begin{tabular}{@{}lcccc@{}}
\toprule
 & Train Acc. & Train AUROC & Test Acc. & Test AUROC\\
\midrule
Before edit & 0.035 & 0.594 & 0.633 & 0.726\\
After edit & 0.640 & 0.553 & 0.627 & 0.709\\
\bottomrule
\end{tabular}
\end{table}

\subsection{Training-seed variation}
\label{app:seed-variation}

Table~\ref{tab:seed-stage12} reports independent training runs for
Vanilla K-SFT (three seeds) and OP-SFT + K$\Delta$-FT (four seeds).
The Vanilla K-SFT seed runs train a LoRA adapter (rank 16, alpha 32) on
standard SFT data built from the update questions, with learning rate
$10^{-4}$, 5 epochs, per-device batch size 8, and gradient accumulation 1.
Table~\ref{tab:seed-fgpo} varies the FGPO seed from one shared
Stage-1+2 initialization. We report the arithmetic mean and sample
standard deviation, $s=\sqrt{\sum_i(x_i-\bar{x})^2/(n-1)}$, over
the available runs; Vanilla K-SFT statistics use its three seeds (43, 44, and 0). These
statistics quantify training-seed variation under the fixed evaluation
split.

\begin{table}[H]
\centering
\caption{Training-seed variation of Vanilla K-SFT and OP-SFT + K$\Delta$-FT
on held-out TriviaQA.}
\label{tab:seed-stage12}
\begin{tabular}{@{}lcccc@{}}
\toprule
 & \multicolumn{2}{c}{Vanilla K-SFT} & \multicolumn{2}{c}{OP-SFT + K$\Delta$-FT}\\
\cmidrule(lr){2-3}\cmidrule(lr){4-5}
Seed & Acc. & AUROC & Acc. & AUROC\\
\midrule
42 & --- & --- & 0.699 & 0.770\\
43 & 0.522 & 0.635 & 0.665 & 0.770\\
44 & 0.522 & 0.627 & 0.670 & 0.790\\
0 & 0.520 & 0.619 & 0.665 & 0.778\\
\midrule
Mean $\pm$ SD & $0.521\pm0.001$ & $0.627\pm0.008$ & $0.675\pm0.016$ & $0.777\pm0.009$\\
\bottomrule
\end{tabular}
\end{table}

\begin{table}[H]
\centering
\caption{FGPO seed variation from one shared Stage-1+2 initialization, evaluated end to end on held-out TriviaQA. Seed 1 is the original full-method run.}
\label{tab:seed-fgpo}
\begin{tabular}{@{}lcc@{}}
\toprule
State & Test Acc. & Test AUROC\\
\midrule
Shared Stage-1+2 initialization & 0.699 & 0.770\\
FGPO, seed 1 & 0.692 & 0.809\\
FGPO, seed 43 & 0.677 & 0.810\\
FGPO, seed 44 & 0.672 & 0.815\\
\midrule
FGPO mean $\pm$ SD & $0.680\pm0.010$ & $0.811\pm0.003$\\
\bottomrule
\end{tabular}
\end{table}

\subsection{Functional FoK and accessible cues}
\label{app:functional-cues}

These analyses measure behavior of the trained Qwen2.5-7B FoK channel.

\paragraph{Consistency at fixed incorrectness.}
From a 5,000-question TriviaQA pool, we sample 16 answers per
question and retain the 1,149 questions where all 16 answers are wrong.
Consistency is the fraction giving the most frequent incorrect answer.
FoK rises with this consistency even though sampled correctness is
zero (Table~\ref{tab:incorrect-consistency}). The reported Spearman
correlation is $\rho=0.24$ ($p=1.8\times10^{-16}$).

\begin{table}[H]
\centering
\caption{Functional FoK among 1,149 questions with 16/16 incorrect sampled answers. Bin labels follow the evaluation grouping.}
\label{tab:incorrect-consistency}
\begin{tabular}{@{}lcc@{}}
\toprule
Answer consistency & Questions & Mean $J$\\
\midrule
0--0.2 & 840 & 0.413\\
0.2--0.4 & 178 & 0.486\\
0.4--0.6 & 71 & 0.571\\
0.6--0.8 & 29 & 0.591\\
0.8--1.0 & 31 & 0.808\\
\bottomrule
\end{tabular}
\end{table}

\paragraph{Accessible reasoning.}
We progressively reveal the generated reasoning trace on 440 MATH
problems using the reasoning-prefix prompts of the preliminary study.
Table~\ref{tab:reasoning-accessibility} shows better discrimination
as more of the trace becomes available. The mean-$J$ gap is the
difference between correct and incorrect solution groups.

\begin{table}[H]
\centering
\caption{FoK discrimination with progressively revealed reasoning on 440 MATH problems.}
\label{tab:reasoning-accessibility}
\begin{tabular}{@{}lcc@{}}
\toprule
Reasoning shown & AUROC & Mean-$J$ gap\\
\midrule
Question only & 0.60 & 0.05\\
25\% & 0.66 & 0.09\\
50\% & 0.67 & 0.10\\
75\% & 0.74 & 0.15\\
Full reasoning & 0.91 & 0.32\\
\bottomrule
\end{tabular}
\end{table}

\paragraph{Context intervention.}
We prepend short contexts to a filtered 1,460-question subset of the
same pool, removing contexts containing a gold
answer. Table~\ref{tab:context-cues} reports FoK and answer accuracy;
deltas are relative to the bare question. Related versus unrelated
words raise mean $J$ by 0.490 (paired Wilcoxon test on per-question
$J$, reported $p=5\times10^{-239}$). Incorrect related sentences
increase $J$ while slightly decreasing accuracy, showing that the
judgment is not determined by correctness alone.

\begin{table}[H]
\centering
\caption{Context interventions on 1,460 questions after filtering gold-answer-containing contexts.}
\label{tab:context-cues}
\begin{tabular}{@{}llcccc@{}}
\toprule
Format & Condition & $J$ & $\Delta J$ & Acc. & $\Delta$Acc.\\
\midrule
--- & None & 0.599 & --- & 0.408 & ---\\
Words & Unrelated & 0.184 & $-0.415$ & 0.424 & $+0.016$\\
Words & Related & 0.674 & $+0.075$ & 0.478 & $+0.070$\\
Sentences & Unrelated & 0.298 & $-0.301$ & 0.436 & $+0.028$\\
Sentences & Correct related & 0.677 & $+0.078$ & 0.510 & $+0.102$\\
Sentences & Incorrect related & 0.772 & $+0.173$ & 0.397 & $-0.011$\\
\bottomrule
\end{tabular}
\end{table}

\subsection{Prompt sensitivity and alternative knowledge updates}

\paragraph{Judgment templates.}
We repeat causal tracing on the same 40 Qwen2.5-7B samples under six
judgment templates (Table~\ref{tab:template-sensitivity}). Recovery
peaks stay at layers 20--21 of 28, with sensitive ranges within layers
16--23. This supports robustness across the tested templates. The templates cover both rephrasings and label
changes; the sensitive range denotes the layers with strong recovery.

\begin{table}[H]
\centering
\caption{Causal-tracing sensitivity to judgment templates on 40 shared samples. Layers out of 28.}
\label{tab:template-sensitivity}
\begin{tabular}{@{}lcc@{}}
\toprule
Template & Peak layer & Sensitive range\\
\midrule
Original & 20 & 16--22\\
A: rephrased & 21 & 17--22\\
B: third-person elicitation & 21 & 16--21\\
C: numeric labels 1/2 & 21 & 18--23\\
D: letter labels A/B & 21 & 18--21\\
E: labels can/cannot & 20 & 18--20\\
\bottomrule
\end{tabular}
\end{table}

\paragraph{Correction strategies.}
Table~\ref{tab:correction-strategies} compares knowledge-update
strategies on Qwen2.5-7B-Instruct using 5,000 TriviaQA update questions
and 1,000 held-out questions from a split different from the main
tables. General capability is measured on the 1,319-question GSM8K
test set~\citep{cobbe2021training}. Reptile uses first-order
meta-updates~\citep{nichol2018reptile}; SDFT uses the released
self-distillation setup~\citep{shenfeld2026sdft}. The lookahead
variant penalizes deviation from Base after a one-step SFT update;
the proximal variant adds an L2 constraint to Base. Average is the
unweighted mean of the three reported accuracies and is given only
when all three are available. Span replacement has the highest average but trails Base on GSM8K.

\begin{table}[H]
\centering
\caption{Alternative knowledge-update strategies on a separate TriviaQA split. Unlikelihood was evaluated on TriviaQA only, so its GSM8K and Average cells show ---.}
\label{tab:correction-strategies}
\begin{tabular}{@{}lcccc@{}}
\toprule
Strategy & Update Acc. & Held-out Acc. & GSM8K & Average\\
\midrule
Base & 0.685 & 0.603 & 0.798 & 0.695\\
Vanilla SFT & 0.992 & 0.452 & 0.368 & 0.604\\
Reptile meta-update & 0.558 & 0.584 & 0.640 & 0.594\\
SDFT & 0.529 & 0.565 & 0.367 & 0.487\\
Unlikelihood on wrong span & 0.500 & 0.528 & --- & ---\\
One-step lookahead KL & 0.738 & 0.619 & 0.646 & 0.668\\
L2 proximal constraint & 0.560 & 0.587 & 0.639 & 0.595\\
Replace wrong span & 0.832 & 0.626 & 0.653 & 0.704\\
\bottomrule
\end{tabular}
\end{table}

\paragraph{Supplementary assets.}
Table~\ref{tab:supplementary-assets} records publisher-specified terms
for additional datasets and the editing implementation. Dataset terms
are distinguished from repository software licenses, particularly for
Natural Questions.

\begin{table}[H]
\centering
\caption{Additional assets and publisher-specified terms.}
\label{tab:supplementary-assets}
\begin{tabular}{@{}lll@{}}
\toprule
Asset & Role & License or terms\\
\midrule
SelfAware & Zero-shot QA & CC BY-SA 4.0 (\href{https://github.com/yinzhangyue/SelfAware}{source})\\
Natural Questions & Transfer and replication & CC BY-SA 3.0 (\href{https://ai.google.com/research/NaturalQuestions/download}{data source})\\
WebQuestions & Zero-shot QA & CC BY 4.0 (\href{https://nlp.stanford.edu/software/sempre/}{source})\\
MATH & Reasoning evaluation & MIT (\href{https://github.com/hendrycks/math/blob/main/LICENSE}{repository})\\
GSM8K & Capability evaluation & MIT (\href{https://github.com/openai/grade-school-math/blob/master/LICENSE}{repository})\\
MEMIT & Editing implementation & MIT (\href{https://github.com/kmeng01/memit/blob/main/LICENSE}{repository})\\
\bottomrule
\end{tabular}
\end{table}

\section{Detailed results for the main-text tables}
\label{app:detailed-main-tables}

\subsection{Knowledge and judgment SFT}
\label{app:detailed-preliminary}

These tables retain all the error-rate columns of
Table~\ref{tab:f2-knowledge-judgment-coupling}. Accuracy, illusion,
and omission are percentages; their changes are percentage points.
AUROC and its changes are unitless. Illusion and omission are conditional
error rates, rather than fractions of all evaluated questions.

\begin{table}[H]
\caption{Base values for the preliminary SFT comparison.}
\label{tab:preliminary-base-full}
\centering
\begin{tabular}{@{}lcccc@{}}
\toprule
Model & Acc. (\%) & AUROC & Illusion (\%) & Omission (\%)\\
\midrule
Qwen2.5-0.5B & 20 & 0.519 & 28 & 85 \\
Qwen2.5-1.5B & 35 & 0.529 & 56 & 38 \\
Qwen2.5-3B & 39 & 0.739 & 8 & 68 \\
Qwen2.5-7B & 53 & 0.724 & 51 & 20 \\
Qwen2.5-14B & 66 & 0.762 & 50 & 19 \\
Qwen2.5-32B & 68 & 0.819 & 62 & 8 \\
LLaMA-3.1-8B & 74 & 0.672 & 81 & 16 \\
\bottomrule
\end{tabular}
\end{table}

\begin{table}[H]
\caption{Changes from Base under K-SFT, J-SFT, and sequential K$\to$J SFT.
Positive accuracy and AUROC changes and negative illusion and omission
changes indicate improvement.}
\label{tab:preliminary-deltas-full}
\centering
\resizebox{\textwidth}{!}{%
\begin{tabular}{@{}lcccccccccccc@{}}
\toprule
 & \multicolumn{4}{c}{K-SFT} & \multicolumn{4}{c}{J-SFT} & \multicolumn{4}{c}{K$\to$J SFT}\\
\cmidrule(lr){2-5}\cmidrule(lr){6-9}\cmidrule(lr){10-13}
Model & Acc. & AUC & Illu. & Om. & Acc. & AUC & Illu. & Om. & Acc. & AUC & Illu. & Om.\\
\midrule
Qwen2.5-0.5B & +2 & -0.016 & -27 & +13 & -14 & +0.213 & -28 & +15 & +2 & +0.211 & -21 & -10 \\
Qwen2.5-1.5B & +6 & -0.035 & +6 & -2 & +2 & +0.255 & -28 & -6 & +5 & +0.248 & -21 & -12 \\
Qwen2.5-3B & +12 & -0.220 & +10 & +18 & +3 & +0.038 & +11 & -29 & +11 & +0.051 & +20 & -41 \\
Qwen2.5-7B & +7 & -0.042 & -15 & +19 & +2 & +0.088 & -9 & -4 & +7 & +0.100 & +6 & -12 \\
Qwen2.5-14B & +6 & +0.004 & +9 & -4 & +1 & +0.030 & +9 & -9 & +5 & +0.068 & +29 & -17 \\
Qwen2.5-32B & +4 & -0.026 & +11 & 0 & 0 & +0.010 & -10 & +2 & +4 & +0.020 & +3 & -3 \\
LLaMA-3.1-8B & +2 & -0.038 & -11 & +9 & +1 & +0.045 & +7 & -11 & +2 & +0.063 & +14 & -14 \\
\bottomrule
\end{tabular}
}
\end{table}

\subsection{Update-set stage-wise ablation}
\label{app:ablation-update-set}

These are the update-set columns of the same runs used in
Table~\ref{tab:ablation}.

\begin{table}[H]
\caption{Update-set results for Base and the three cumulative training stages.}
\label{tab:ablation-update-set}
\centering
\begin{tabular}{@{}lccccc@{}}
\toprule
Stage & Acc.$\uparrow$ & AUROC$\uparrow$ & Illu.$\downarrow$ & Om.$\downarrow$ & Bias$\to0$\\
\midrule
Base & 0.680 & 0.702 & 0.653 & 0.109 & 0.381 \\
Stage 1: + OP-SFT & 0.966 & 0.575 & 0.589 & 0.319 & 0.454 \\
Stage 2: + K$\Delta$-FT & 0.967 & 0.803 & 0.350 & 0.190 & 0.270 \\
Stage 3: + FGPO & 0.961 & 0.817 & 0.250 & 0.240 & 0.245 \\
\bottomrule
\end{tabular}
\end{table}

\end{document}